\pdfoutput=1
\documentclass[11pt]{article}

\usepackage[utf8]{inputenc}
\usepackage[T1]{fontenc}
\usepackage{lmodern}
\usepackage[a4paper,margin=1in]{geometry}

\usepackage{amsmath,amssymb,amsfonts}
\usepackage{bm}
\usepackage{textcomp}
\usepackage{pifont}

\usepackage{graphicx}
\usepackage{booktabs}
\usepackage{makecell}
\usepackage{multirow}
\usepackage{array}
\usepackage{float}
\usepackage{subfig}
\usepackage{algorithm}
\usepackage{algorithmic}

\usepackage{enumitem}
\usepackage{mdframed}
\usepackage{xcolor}

\usepackage{authblk}
\usepackage[hidelinks]{hyperref}

\usepackage[sorting=none,backend=biber]{biblatex}
\definecolor{accessgreen}{rgb}{0.3, 0.7, 0.3}
\definecolor{accessred}{rgb}{0.8, 0.2, 0.2}
\definecolor{accessorange}{rgb}{1.0, 0.6, 0.0}

\DeclareRobustCommand{\cmark}{%
  \mbox{%
    \ooalign{%
      \hfil\textcolor{accessgreen!20}{\fontsize{24}{0}\selectfont\textbullet}\hfil\cr
      \hfil\raisebox{1ex}{\textcolor{accessgreen}{\footnotesize\ding{51}}}\hfil\cr
    }%
  }%
}

\DeclareRobustCommand{\xmark}{%
  \mbox{%
    \ooalign{%
      \hfil\textcolor{accessred!20}{\fontsize{24}{0}\selectfont\textbullet}\hfil\cr
      \hfil\raisebox{1ex}{\textcolor{accessred}{\footnotesize\ding{55}}}\hfil\cr
    }%
  }%
}

\DeclareRobustCommand{\imark}{%
  \mbox{%
    \ooalign{%
      \hfil\textcolor{accessorange!30}{\fontsize{24}{0}\selectfont\textbullet}\hfil\cr
      \hfil\raisebox{1ex}{\small$\sim$}\hfil\cr
    }%
  }%
}



\newcommand{\keywords}[1]{%
  \vspace{0.5em}
  \noindent\textbf{Keywords:} #1
}

\title{AIDEN: Design and Pilot Study of an AI Assistant for the Visually Impaired}

\author[1]{Luis Marquez-Carpintero\thanks{Corresponding author: Luis Marquez-Carpintero, e-mail: \texttt{luis.marquez@ua.es}. This work was fully supported by Indra and Fundación Universia.}}
\author[1]{Francisco Gomez-Donoso}
\author[2]{Zuria Bauer}
\author[1]{Bessie Dominguez-Dager}
\author[1]{Alvaro Belmonte-Baeza}
\author[1]{Mónica Pina-Navarro}
\author[1]{Francisco Morillas-Espejo}
\author[1]{Felix Escalona}
\author[1]{Miguel Cazorla}

\affil[1]{Institute for Computer Research, University of Alicante, Alicante, Spain}
\affil[2]{ETH Zurich, Zurich, Switzerland}

\date{}

\begin{document}

\maketitle

\begin{abstract}
This paper presents AIDEN, an artificial intelligence--based assistant designed to enhance the autonomy and daily quality of life of visually impaired individuals, who often struggle with object identification, text reading, and navigation in unfamiliar environments. Despite these advancements, a significant research gap remains: there is a lack of integrated solutions that provide real-time, actionable guidance without compromising the user's environmental awareness or data privacy. Current systems force a trade-off between functional assistance and sensory safety. AIDEN addresses these limitations through a novel system integration that combines real-time object detection and object description within a multimodal interaction framework, prioritizing the reduction of auditory overload through tactile feedback. A central design feature is that AIDEN offloads spatial guidance to continuous haptic feedback for object retrieval, while audio is primarily used for semantic content, including OCR and VQA. Privacy is preserved by ensuring that no personal data are stored. Empirical evaluations with visually impaired participants assessed perceived ease of use and acceptance using the Technology Acceptance Model (TAM). Results indicate high user satisfaction, particularly regarding intuitiveness and perceived autonomy. Moreover, the ``Find an Object'' module achieved low-latency performance under the tested conditions. These findings suggest that multimodal haptic--visual feedback may support daily usability and independence relative to traditional audio-centric methods, motivating larger-scale validations.
\end{abstract}


\section{Introduction}
\label{sec:intro}
{
One of the most promising applications of artificial intelligence (AI) is the development of technologies to support people with impairments. In particular, AI-based assistants for visually impaired users may support quality of life by increasing autonomy in everyday tasks and enabling more independent interaction with the surrounding environment.
Visually impaired individuals face recurring challenges in daily life, including identifying objects, understanding visual scenes, and navigating unfamiliar environments. Tasks such as locating items in a supermarket or interpreting informational signs can pose substantial difficulties due to the lack of visual input~\cite{manduchi2012mobile}, which can reduce autonomy and negatively affect daily functioning.
Several solutions exist to support these needs. Screen readers and audiobooks facilitate access to written content. However, many of these approaches rely primarily on audio feedback. In practice, this can be problematic in noisy environments~\cite{ahmetovic2023enhancing}, can contribute to auditory overload~\cite{zhao2024hearing}, and must be carefully designed to avoid masking critical environmental cues~\cite{theodorou2019developing}. { AIDEN addresses these limitations by offloading spatial guidance to continuous haptic feedback, thereby reserving the auditory channel for essential semantic content like OCR or VQA. This multimodal balance ensures that users maintain situational awareness in unpredictable daily contexts where audio cues might otherwise be masked.}

Beyond audio-related issues, current assistive technologies face additional practical limitations that hinder sustained real-world adoption. First, many tools remain \emph{fragmented}, each focused on a single function (e.g., Optical Character Recognition, OCR, or Object Description), forcing users to switch between applications and interaction styles~\cite{granquist2021evaluation}.

Second, several solutions rely on cloud services or third-party assistance, raising privacy and data-control concerns because users cannot visually verify what is being captured or transmitted, and bystanders may be unintentionally recorded~\cite{ahmed2015privacy}. Finally, beyond recognizing \emph{what} is present, users often need actionable guidance about \emph{where} an object is in order to physically retrieve it, which is not well supported by purely verbal feedback in dynamic environments. {AIDEN addresses these limitations by unifying key assistive functions within a single workflow and implementing a real-time, closed-loop ``Find an Object'' module. This feature provides actionable spatial guidance for precise camera aiming and object localization.}

{ Privacy and data control represent another critical barrier to adoption. Several solutions rely on cloud services or third-party assistance where users cannot visually verify the camera's field of view, raising concerns about inadvertently capturing sensitive information or bystanders. To bridge this privacy gap, AIDEN adopts a data-minimization approach where no personal data are stored, this guarantees a transparent and private data management.}

{ Finally, some systems require specialized hardware, such as head-mounted displays, which reduces affordability, limits scalability, and potentially increases social stigma. Additionally, the high power consumption of such devices often prevents the all-day battery life that users enjoy with standard smartphones. AIDEN prioritizes accessibility through a smartphone-first architecture that leverages the social acceptability of mainstream devices. Moreover, using a distributed client-server pipeline allows the system to perform complex AI tasks on resource-constrained hardware without the need for specialized equipment.}

Together, these limitations motivate a unified, smartphone-based assistant that complements speech with an alternative guidance channel, integrates key assistive functions within a single workflow, and follows a privacy-preserving data-minimization approach. We therefore propose \textbf{AIDEN}, a multimodal assistant that combines \textit{OCR}, \textit{Object Description / visual question answering (VQA)}, and \textit{real-time object finding}.}

{Specifically,} this paper makes three main contributions:
\begin{itemize}
    \item \textbf{Accessibility-first multimodal interaction:} We introduce a practical interaction paradigm that reduces auditory overload by combining screen-reader-compatible navigation with continuous haptic guidance for object retrieval.
    \item \textbf{End-to-end system integration:} We present a cross-platform mobile application that unifies OCR, scene understanding (Object Description/VQA), and object finding within a smartphone-first client--server pipeline optimized for resource-constrained devices and guided by data minimization.
    \item \textbf{Empirical evidence with visually impaired users:} We report results from a pilot evaluation with visually impaired participants, including Technology Acceptance Model (TAM) usability outcomes and technical runtime performance measurements.
\end{itemize}

\section{Literature Review}
\label{sec:lit_review}

{
To position AIDEN within the current landscape, we first analyze the evolution of environmental understanding capabilities. This is followed by an examination of interaction channels and the challenges of auditory overload. Finally, we address the practical barriers, such as hardware costs and privacy concerns, that frequently hinder the real-world adoption of these tools.
}

\subsection{AI-Based Assistive Applications and Hardware Constraints}
\label{sub:lit_rev_a}
{
Modern assistive applications leverage deep learning and computer vision to support tasks such as text reading, object identification, and scene understanding~\cite{kisanga2025enhancing, adam2025leveraging}. Commercial tools including Seeing AI~\cite{SeeingAI2024}, Be My Eyes~\cite{BeMyEyes2024}, and Envision AI~\cite{envision2026} demonstrate the maturity of these capabilities in real-world products. However, qualitative evidence suggests that users continue to prioritize mobility-related assistance, with navigation frequently reported as a dominant unmet need~\cite{moon2022factors}.
}
{
Research prototypes also explore richer forms of guidance using head-mounted mixed reality. For example, MR.NAVI~\cite{mrnavi2025} uses a HoloLens 2 headset to deliver spatial guidance based on real-time scene understanding. While this line of work highlights the potential of mixed reality devices for context-aware assistance, reliance on expensive and power-demanding hardware can limit affordability and broad adoption.
}
{ In contrast, a smartphone-first approach leverages the ubiquity and social acceptability of mainstream devices, offering a more discrete and cost-effective solution that integrates seamlessly into the user's daily life without the social friction often associated with specialized wearables.}
{ Ultimately, the current landscape suggests that despite technological advancements, important challenges remain in addressing the specific needs of individuals with visual impairments. This population faces recurring challenges in essential tasks such as identifying objects in unfamiliar environments or interpreting informational signs, which directly impacts their autonomy and quality of life. While privacy is not always explicitly addressed in existing solutions, many also rely primarily on auditory feedback, which can lead to auditory overload. AIDEN was designed as an integrated approach tailored to this user population. Our approach prioritizes the preservation of situational awareness by offloading spatial guidance to continuous haptic channels, thereby using the smartphone as an assistive tool that respects the user’s vital reliance on their auditory environment while supporting potentially safer and more private navigation.}

\subsection{Privacy and Data-Control Constraints}
\label{sub:lit_rev_b}
{
Another adoption barrier concerns privacy and data control in camera-based assistants, particularly when processing is performed in the cloud or involves third-party services. Research on human factors highlights a critical privacy gap: since visually impaired users cannot visually verify the camera's precise field of view before data transmission, there is a documented risk of inadvertently sharing sensitive information, such as financial documents, medical records, or bystander imagery. Studies by Ahmed et al.~\cite{ahmed2015privacy} confirm that this lack of visual control and the resulting risk of sharing sensitive content constitute a significant barrier to the adoption of cloud-based assistive tools. As a result, privacy-preserving designs typically emphasize transparency, data minimization, and limiting unnecessary capture or sharing wherever possible.
In order to synthesize how AIDEN addresses the various research gaps identified, ranging from auditory overload and hardware constraints to the aforementioned privacy concerns, Table~\ref{tab:comparison_final_transposed} provides a comparative analysis against representative commercial assistive technologies. The comparison highlights why existing solutions often may involve trade-offs between functionality and sensory load. The selected systems—\textit{Seeing AI}, \textit{Be My Eyes}, \textit{Envision AI}, and \textit{Meta Ray-Ban} were chosen as they represent the current state-of-the-art in smartphone-based and wearable assistance. The comparison criteria were derived directly from the primary research gaps: the reduction of auditory overload, the requirement for active spatial guidance, data privacy, advanced semantic understanding (VQA), and the goal of maintaining high accessibility through smartphone-only deployment. The methodology used to build Table~\ref{tab:comparison_final_transposed} was a hands-on, feature-by-feature comparative evaluation of the selected systems. For each criterion, we tested the apps/devices in practice to verify whether the functionality was available and how it behaved, and we cross-checked our observations against official documentation.
}

\begin{table*}[ht]
\centering
\renewcommand{\arraystretch}{1.5}
\setlength{\tabcolsep}{3pt}
\begin{tabular}{
    >{\raggedright\arraybackslash}m{2.7cm}  
    >{\centering\arraybackslash}m{1.2cm}    
    >{\centering\arraybackslash}m{1.2cm}    
    >{\centering\arraybackslash}m{2.0cm}    
    >{\centering\arraybackslash}m{2.5cm}    
    >{\centering\arraybackslash}m{1.5cm}    
    >{\centering\arraybackslash}m{1.8cm}    
}
\hline
\textbf{System} & 
\multicolumn{2}{c}{\textbf{Guidance Feedback}} & 
\textbf{Object Search} & 
\textbf{Privacy \& Data} & 
\textbf{VQA Capability} & 
\textbf{Smartphone Only} \\ 
\cline{2-3}
& \textbf{Haptic} & \textbf{Audio} & & & & \\
\hline
\textbf{Seeing AI}~\cite{SeeingAI2024} & 
\raisebox{-1ex}{\xmark} & 
\raisebox{-1ex}{\cmark} & 
\makecell{
    \raisebox{-1ex}{\imark} \\ 
    \scriptsize (Post-photo and \\ \scriptsize finger interaction)} 

& \makecell{
    \raisebox{-1ex}{\imark} \\ 
    {\scriptsize (Privacy under} \\ \scriptsize {subscription)}} &
    
\raisebox{-1ex}{\xmark} & 
\raisebox{-1ex}{\cmark} \\

\hline
\textbf{Be My Eyes}~\cite{BeMyEyes2024} & 
\raisebox{-1ex}{\xmark} & 
\raisebox{-1ex}{\cmark} & 
\raisebox{-1ex}{\xmark} & 
\raisebox{-1ex}{\xmark} & 
\raisebox{-1ex}{\cmark} & 
\raisebox{-1ex}{\cmark} \\
\hline
\textbf{Envision AI}~\cite{envision2026} & 
\raisebox{-1ex}{\xmark} & 
\raisebox{-1ex}{\cmark} & 
\makecell{\raisebox{-1ex}{\imark} \\ \scriptsize (Non orientation \\ \scriptsize guide)} &

\makecell{
    \imark \\
    {    
        \scriptsize (Privacy under} \\ {\scriptsize subscription)}
} & 

\vspace{2mm}
\cmark & 
\vspace{2mm}
\cmark \\
\hline
\textbf{Meta Ray-Ban}~\cite{MetaRayBan} & 
\vspace{2mm}
\xmark & 
\vspace{2mm}
\cmark & 
\vspace{2mm}
\xmark & 
\vspace{2mm}
\xmark & 
\vspace{2mm}
\cmark & 
\vspace{2mm}
\xmark \\
\hline
\textbf{AIDEN (Ours)} & 
\vspace{2mm}
\cmark & 
\vspace{2mm}
\cmark & 
\vspace{2mm}
\cmark & 
\vspace{2mm}
\cmark & 
\vspace{2mm}
\cmark & 
\vspace{2mm}
\cmark \\
\hline
\end{tabular}
\caption{
    {Comparative analysis: AIDEN vs. leading commercial assistive technologies. Criteria are based on identified research gaps in auditory overload and privacy. Symbols represent: full integration (\raisebox{-1ex}{\cmark}), not available (\raisebox{-1ex}{\xmark}), and partial or conditional support/subscription-gated (\raisebox{-1ex}{\imark}). Data obtained through hands-on testing and official documentation review.}
}

\label{tab:comparison_final_transposed}
\end{table*}

\subsection{Interaction Paradigms and Auditory Load}
{
A key design dimension is how assistive systems deliver guidance. As shown in our comparative evaluation of leading commercial systems (Table~\ref{tab:comparison_final_transposed}), current market leaders primarily rely on auditory feedback loops, such as Text-To-Speech (TTS) and sonification, for user guidance. This reliance on the auditory channel is well documented in the literature as a potential source of cognitive overload and masked environmental cues. Prior work has shown that continuous audio can be challenging in noisy contexts~\cite{ahmetovic2023enhancing} and may contribute to auditory overload~\cite{zhao2024hearing}. Moreover, auditory interfaces must be designed carefully to avoid masking environmental sounds that are critical for safety and situational awareness~\cite{theodorou2019developing, stevens2005auditory}. These findings motivate interaction paradigms that reduce dependence on continuous speech for spatial guidance by assigning part of the guidance task to alternative sensory channels.
}
{
In this context, recent research has explored haptic feedback as a means to mitigate auditory overload. A systematic review of 32 papers~\cite{xia2021haptic} identified three dominant haptic guidance strategies: vibrotactile, kinesthetic, and shape-changing modalities. Shape-changing interfaces~\cite{quinn2024shape} have demonstrated superior speed in localizing three-dimensional targets compared with single-axis vibration, while also avoiding tactile adaptation associated with prolonged vibratory stimulation. Likewise, the ALVU wearable system~\cite{katzschmann2018safe} validated portable navigation with a sensor-to-haptic latency of approximately 120\,ms, achieving task completion rates comparable to white cane performance in blind users. Together, these findings suggest that strategically assigning spatial information to haptic channels can preserve both user autonomy and situational awareness, while reducing reliance on continuous auditory guidance.
This rationale also helps interpret the comparative results in Table~\ref{tab:comparison_final_transposed}. In the table, the symbol $\sim$ indicates partial or conditional support rather than full integration. For Seeing AI, object search was marked as partial because the tested interaction relied on post-photo and finger-based interaction, rather than closed-loop, real-time guidance for centering and retrieving a target object. Its privacy/data criterion was also marked as partial because the corresponding support was considered available only under subscription. For Envision AI, object search was marked as partial because the available functionality acted as a non-orientation guide rather than a full real-time closed-loop localization mechanism. Its privacy/data criterion was likewise marked as partial because this support was considered subscription-dependent.}

\subsection{User Requirements and Multimodal Support Systems}
{
The autonomy and quality of life of visually impaired individuals are often hindered by difficulties in object identification, text reading, and navigation in unfamiliar environments. Beyond environmental recognition, users frequently require actionable guidance to determine an object's spatial location for physical retrieval. To address these needs, prior research has explored various sensory modalities beyond purely visual or mixed reality interfaces. While screen readers and audiobooks facilitate access to written content, the reliance on the auditory channel can lead to sensory overload. Consequently, alternative modalities such as haptic and tactile feedback have been explored to offload spatial guidance, allowing the auditory channel to be reserved for semantic content like OCR or VQA. For users with low vision, support systems also integrate high-contrast color schemes and scalable typography to maximize residual vision. The combination of auditory cues, visual highlights, and haptic signals forms a multimodal triad that enhances confirmation of actions and precise navigation.
}

{
As summarized in Table~\ref{tab:comparison_final_transposed}, most commercial assistants remain predominantly audio-driven and provide limited support for closed-loop, real-time object localization. In addition, privacy/data-control guarantees vary across solutions, and smartphone-only deployment is not universal when wearables are required. Overall, the combination of reduced reliance on continuous audio for spatial guidance, real-time closed-loop object finding, privacy-aware operation, and smartphone-only accessibility remains insufficiently addressed by existing systems. This gap motivates the design choices explored in AIDEN.
}

\subsection{Human Factors and Technology Acceptance Models}
\label{sub:lit_rev_c}

{
Beyond technical performance, the effectiveness of assistive applications depends on user acceptance and sustained use. The Technology Acceptance Model (TAM)~\cite{davis1989technology} and extensions such as Unified Theory of Acceptance and Use of Technology (UTAUT) have been applied to evaluate assistive technologies~\cite{djamasbi2006accessibility, bag2016acceptance}. Prior work suggests that, for visually impaired users, adoption can be influenced not only by perceived usefulness and ease of use but also by disability-specific factors such as training, which can reduce barriers and increase behavioral intention~\cite{theodorou2024challenges}. Accordingly, we structure our evaluation of AIDEN using TAM constructs and incorporate training to analyze its impact on adoption-related outcomes.
}

\section{Methodology}
\label{sec:approach}

AIDEN is implemented as a cross-platform mobile application aimed at assisting visually impaired individuals in their daily activities. It integrates {machine learning} and deep learning techniques to enable three core functionalities: ``Text-To-Speech'', ``Object Description'' with VQA, and ``Find an Object''. These features are accessed through a user-friendly interface specifically designed with accessibility in mind, aiming to improve the quality of life of users with visual impairments.

The application is developed for both Android and iOS platforms using a hybrid architecture based on the Ionic framework (v6.20.1), Capacitor Core (v4.7.0), and Vue.js (v3.3.7). This stack facilitates deployment across multiple platforms, including web, while maintaining a consistent user experience. Accessibility support is enhanced through the integration of the Capacitor ScreenReader plugin~\cite{capacitor_screen_reader}, which enables screen content narration, and the Vue i18n library~\cite{vue_i18n}, which provides multilingual support. The following subsections detail the implementation of each of AIDEN’s core functionalities.

\subsection{Participants}

This study involved a cohort of 28 participants ($N=28$), informed consent was obtained from all individual participants included in the study, with visual impairments. The recruitment of this specialized, low-incidence population was facilitated through formal collaboration with the National Organization of Spanish Blind People (ONCE) and the Retina Comunidad Valenciana association. This necessity reflects the resource constraints in conducting assistive technology research in resource-limited environments, where the available evidence is limited in quantity and quality, unevenly distributed across assistive technology types, and largely based on observational study designs~\cite{matter2017assistive}. To ensure an unbiased evaluation, none of the participants were involved in the system's development; their input was collected only after iterative feedback from accessibility experts to assess user acceptance and usability.

As detailed in Table~\ref{tab:participants}, a total of 28 visually impaired individuals (15 males and 13 females) were recruited. To ensure heterogeneity, the cohort included both blind participants and participants with low vision. In our sample, the blind subgroup corresponded to total blindness (no light perception), whereas the low-vision subgroup comprised profound visual impairment and severe visual impairment, reflecting different levels of residual vision.

A crucial aspect of the recruitment was assessing the participants' technological proficiency. All participants possessed at least intermediate experience with smartphones and were familiar with standard accessibility tools, such as screen readers (e.g., TalkBack on Android or VoiceOver on iOS). This methodological step ensured that the evaluation focused on the intrinsic usability of the AIDEN system itself, rather than external factors such as the basic learning curve of using a touchscreen device. {Consequently, the study focuses on users with at least intermediate proficiency to evaluate the system’s specific assistive features rather than basic touchscreen navigation.}

\begin{table}[H]
\centering
\resizebox{\columnwidth}{!}{%
    \begin{tabular}{lcccc}
    \hline\textbf{Variable} & \textbf{Frequency} & \textbf{Percentage} \\ \midrule
    \textbf{Sex} & & \\ \midrule
    Male & 15 & 53.57\% \\
    Female & 13 & 46.43\% \\ \midrule
    \textbf{Age} & & \\ \midrule
    20-30 years & 4 & 14.29\% \\
    31-40 years & 5 & 17.86\% \\
    41-50 years & 9 & 32.14\% \\
    51-60 years & 7 & 25.00\% \\
    60+ years & 3 & 10.71\% \\ \midrule
    \textbf{Degree of Visual Impairment} & & \\ \midrule
    Total Blindness & 14 & 50.00\% \\
    Profound Visual Impairment & 10 & 35.71\% \\
    Severe Visual Impairment & 4 & 14.29\% \\ \midrule
    \textbf{Operating System} & & \\ \midrule
    iOS & 20 & 71.43\% \\
    Android & 8 & 28.57\% \\ 
    \hline
    \end{tabular}%
}
\caption{Demographics and technological profile of participants. Total Blindness, defined by a complete lack of light perception, Profound Visual Impairment, where users retain only light and shadow perception, and Severe Visual Impairment, characterized by functional residual vision that allows for the perception of forms and shapes.}
\label{tab:participants}
\end{table}

{
The study is intentionally positioned as a formative usability pilot with a targeted cohort ($N=28$), aligned with established usability engineering evidence showing diminishing returns in the discovery of new usability issues after a handful of participants (often around five)~\cite{virzi1992refining,nielsen1993mathematical}. Accordingly, our goal was to identify major interaction breakdowns and practical barriers in early-stage validation, rather than to estimate population-level effects.
While the collected data allow us to report qualitative insights and descriptive summaries (such as task completion, runtime, and questionnaire distributions) the sample size remains insufficient for strong inferential or broadly generalizable claims. Although the expanded cohort improves the descriptive value of the study, the statistical analyses reported in this work should be interpreted as exploratory rather than confirmatory.
}

\subsection{Accessible Interaction Design}
\label{sec:interaction_design}

To ensure the system is fully usable by individuals with varying degrees of visual impairment, the user interface was designed following a screen-reader-first design philosophy and aiming to conform to the Web Content Accessibility Guidelines (WCAG) 2.1 Level AA~\cite{wcag21}.

Contrary to standard visual interfaces where interaction relies on hand-eye coordination, AIDEN's interaction model relies on the underlying accessibility tree provided by the operating system. The application is fully integrated with native screen readers, \textit{TalkBack} on Android and \textit{VoiceOver} on iOS. 

Consequently, the interaction flow is defined as follows:

\begin{itemize}
    \item \textbf{Non-visual navigation:} All interactive elements (buttons, menus) are tagged with semantic descriptions (accessibility labels). Users do not need to visually locate a button; instead, they navigate the interface using standard gestures (e.g., linear swiping to move focus, double-tapping to activate) familiar to proficient screen reader users. This negates the need for precise tactile targeting.
    
    \item \textbf{Low vision support:} For participants with residual vision, the GUI implements high-contrast color schemes and scalable typography to maximize legibility, as defined in the configuration settings.
    
    \item \textbf{Multimodal feedback:} To confirm actions (such as capturing an image or sending a request to the AI model), the system employs a triad of feedback mechanisms: auditory cues (TTS), visual highlights (for low vision), and haptic feedback (vibration patterns).
\end{itemize}

{For the "Find an Object" functionality, where the camera must be pointed at a specific target, we implemented a continuous guidance system. To facilitate object localization, we utilized the Geiger-counter metaphor~\cite{korff1958geiger}, a feedback paradigm where the frequency of stimuli increases as a user approaches a target. In its original context, a radiation detector emits a higher rate of 'clicks' when nearing a radioactive source, we adapted this principle into a multimodal feedback mechanism for spatial guidance.In AIDEN’s implementation, the frequency of both vibrations and audio beeps is inversely proportional to the distance between the camera frame's optical center and the detected object's centroid. }

\subsubsection{Real-time Spatial Guidance and Haptic Feedback}
\label{sec:spatial_guidance}

A critical challenge for visually impaired users in object detection tasks is not merely identifying \textit{what} is in the scene, but \textit{where} it is located relative to the camera to facilitate physical grasping. Regarding the mechanism of this interaction, we detail our guidance algorithm below.

The ``Find an Object'' module implements a closed-loop feedback system that guides the user to center the target object (e.g., keys, cup) in the camera frame without visual references. The algorithm operates in two concurrent stages:

\begin{enumerate}
    \item \textbf{Coarse navigation (directional TTS):} 
    When the target object is detected but resides in the periphery of the image (outside the center of the screen), the system calculates the quadrant of the object's bounding box. Based on this position, AIDEN emits specific verbal instructions via the TTS engine, such as: ``Move camera to the left,'' ``Move camera to the right,'' ``Tilt up,'' or ``Tilt down''.

    \item \textbf{Fine positioning (continuous audio--haptic feedback)}:
    Once the object enters the central region, verbal commands become too slow for real-time adjustments. The system switches to a continuous audio--haptic feedback loop and maps the centering error to feedback rate. 
\end{enumerate}

{As formalized in Algorithm~\ref{alg:haptic_guidance}, the guidance logic relies on the Euclidean distance $d$ between the detected object's centroid $O(x,y)$ and the camera's optical center $C(x,y)$. This distance $d$ is inversely mapped to the frequency of haptic pulses (vibrations) and audio beeps. As the user moves the camera and $d$ decreases (the object becomes more centered), the frequency increases to provide high-resolution spatial awareness:}

\begin{algorithm}
\begin{algorithmic}[1]
\REQUIRE Object Bounding Box $B$, Frame Center $C(x,y)$, Threshold $T_{center}$
\ENSURE Haptic Pulses $P$, Audio TTS $A$
\STATE $O(x,y) \leftarrow \text{calculate\_centroid}(B)$
\STATE $d \leftarrow \sqrt{(O_x - C_x)^2 + (O_y - C_y)^2}$ 
\IF{$d > T_{center}$}
    \STATE \COMMENT{Phase 1: Coarse Navigation}
    \STATE $Q \leftarrow \text{determine\_quadrant}(O)$ 
    \STATE $A \leftarrow \text{generate\_TTS\_instruction}(Q)$ 
\ELSE
    \STATE \COMMENT{Phase 2: Fine Positioning}
    \IF{$d$ is Peripheral}
        \STATE $P \leftarrow \{200\text{ms ON, } 300\text{ms OFF}\}$ 
    \ELSIF{$d$ is Approaching}
        \STATE $P \leftarrow \{200\text{ms ON, } 100\text{ms OFF}\}$ 
    \ELSIF{$d \approx 0$ (Centered)}
        \STATE $P \leftarrow \text{Continuous Vibration}$
        \STATE $A \leftarrow \text{Confirmation Tone}$
    \ENDIF
\ENDIF
\RETURN $P, A$
\end{algorithmic}
\caption{Multimodal Geiger-Counter Spatial Guidance}
\label{alg:haptic_guidance}
\end{algorithm}

\begin{itemize}
    \item \textbf{Peripheral detection:} Intermittent vibration pulses (200 ms ON, 300 ms OFF).
    \item \textbf{Approaching center:} Rapid pulsing  (200 ms ON, 100 ms OFF).
    \item \textbf{Locked/Centered:} Continuous steady vibration and a distinct confirmation tone, signaling the user that the object is directly ahead and reachable.
\end{itemize}

{This multimodal approach ensures that the user receives instantaneous spatial information, compensating for the lack of visual depth perception and enabling precise camera aiming.
}

\subsection{Question Design}
\begin{figure}[h]
    \centering
    \subfloat[Home view]{%
        \includegraphics[width=0.22\textwidth]{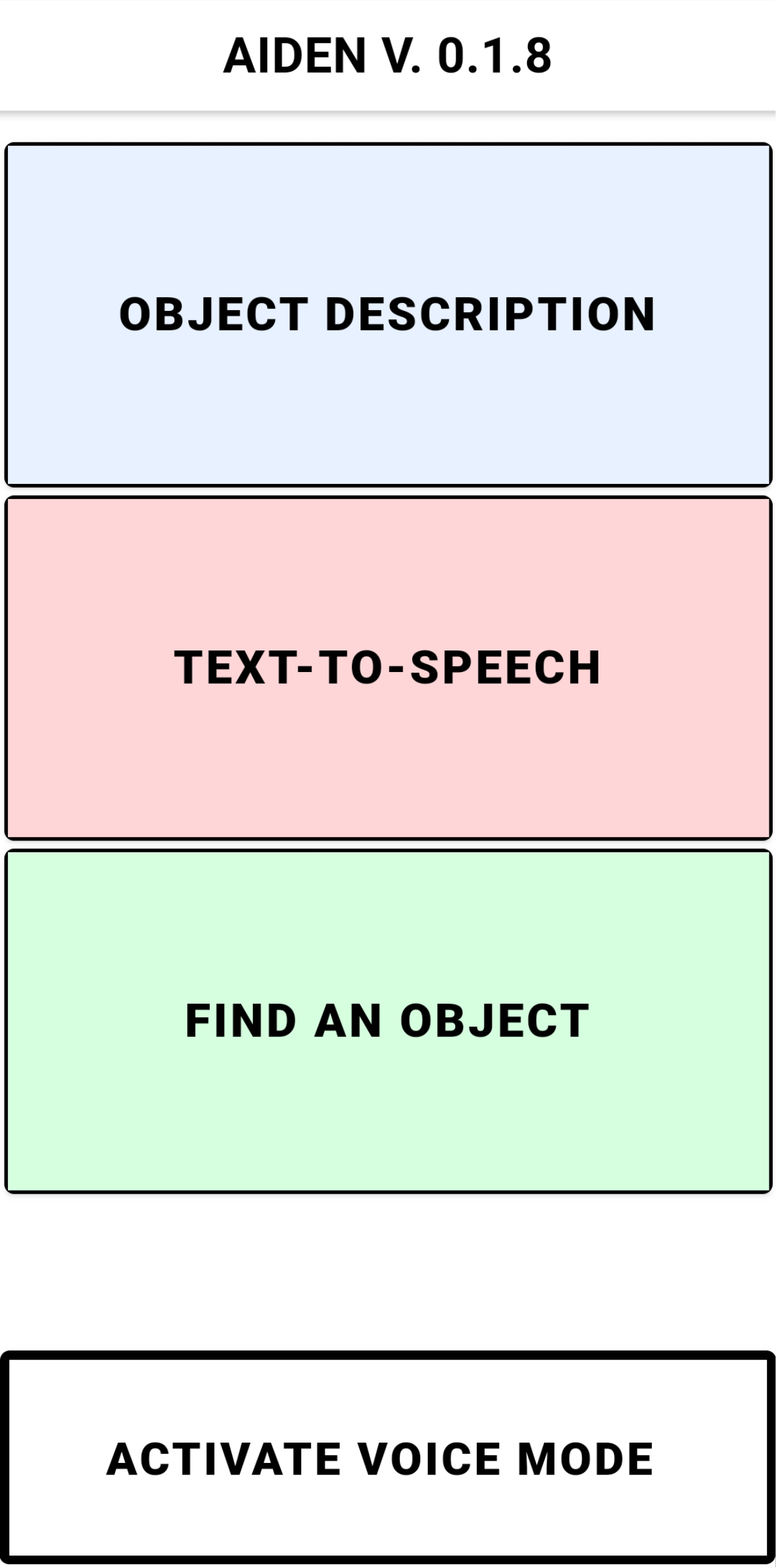}
        \label{fig:home}
    }
    \hfill
    \subfloat[“Object Description” and “Text-To-Speech” view]{%
        \includegraphics[width=0.22\textwidth]{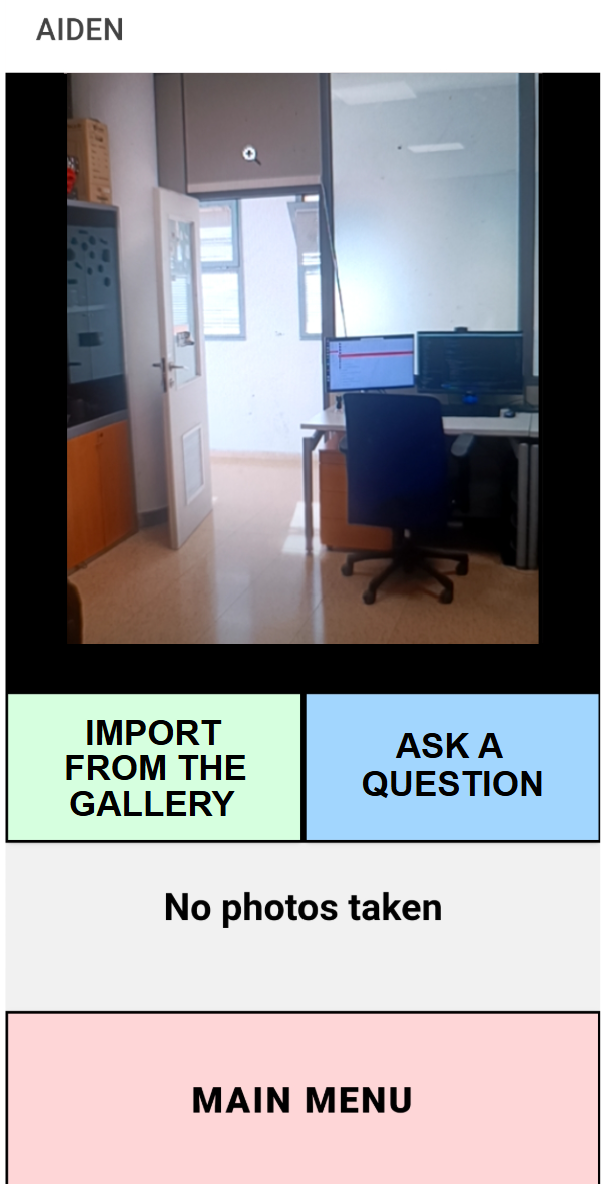}
        \label{fig:camera_view}
    }
    \hfill
    \subfloat[Entering information via text]{%
        \includegraphics[width=0.22\textwidth]{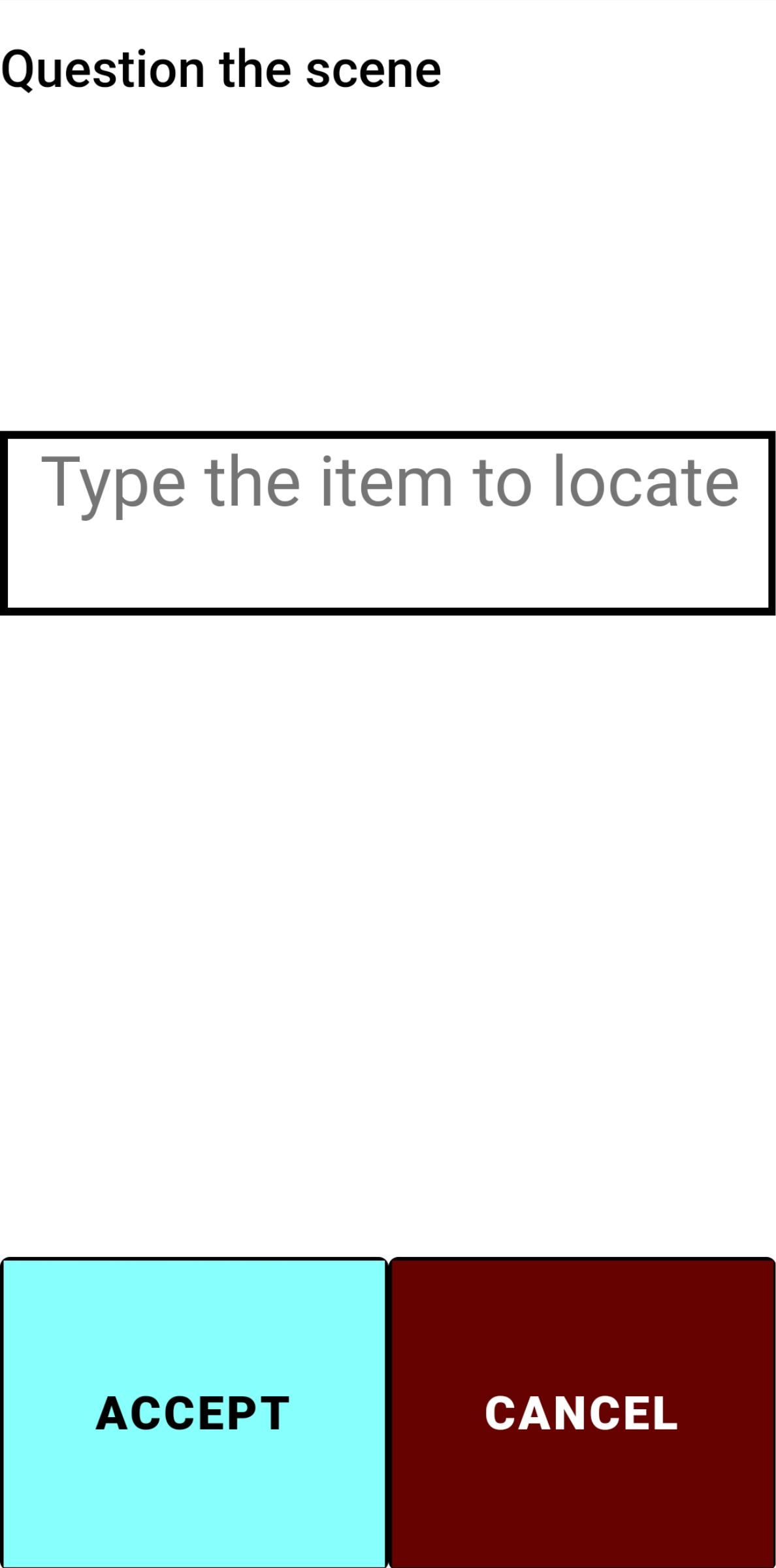}
        \label{fig:modal}
    }
    \hfill
    \subfloat[View to locate an element in the space]{%
        \includegraphics[width=0.22\textwidth]{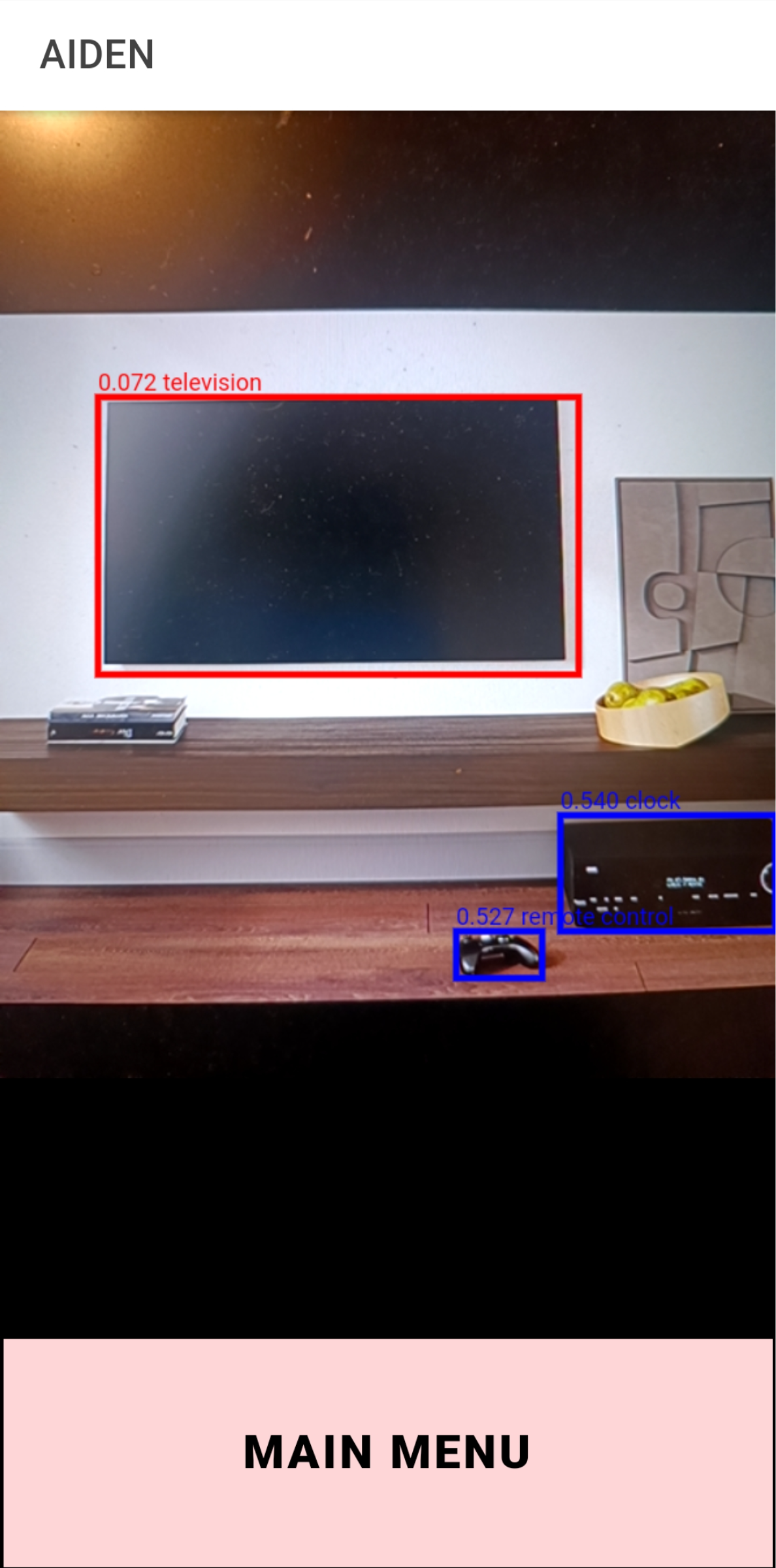}
        \label{fig:location}
    }

    \caption{User interface views of the application.}
    \label{fig:ui_views}
\end{figure}

The user interface is organized into two main views: the home screen, which provides access to the three core functionalities (``Object Description'', ``Text-To-Speech'', and ``Find an Object''), and the capture view, where users can either take a new photo or import an existing one to receive model-generated feedback. Figure \ref{fig:ui_views} illustrates the key interaction screens of the application.

To assess user experience and acceptance, a questionnaire was administered consisting of the following items:

\begin{enumerate}[label=Q\arabic*:]
    \item I found the system sufficiently intuitive for practical use.
    \item I believe the system supports autonomous usage by the user.
    \item I believe visually impaired users will find the assistance system enjoyable to use.
    \item I found the proposed system to be highly motivating for users.
    \item I believe using this system for daily tasks is beneficial.
    \item I consider the app to be a valuable information resource.
    \item I found the object search and text reading features to be effective for personal independence.
    \item I found the prototype to be well-conceived.
    \item I believe the system's design is appealing to the user.
    \item I believe the system's features are attractive to prospective users.
    \item The prototype's features are straightforward and comprehensible enough for widespread use.
    \item The proposal adequately addresses the need for a support system for visually impaired people.
    \item I would be interested in utilizing the system in a real-world scenario and implementing it in practice.
    \item I would like to incorporate the final version of the system into my daily life and routines.
    \item I would recommend employing the final version of the system for user interactions.
    \item Overall, I rate the proposed system positively.
\end{enumerate}

\subsection{Materials and Equipment}

\begin{figure*}[ht]
    \centering
    \includegraphics[width=0.7\textwidth]{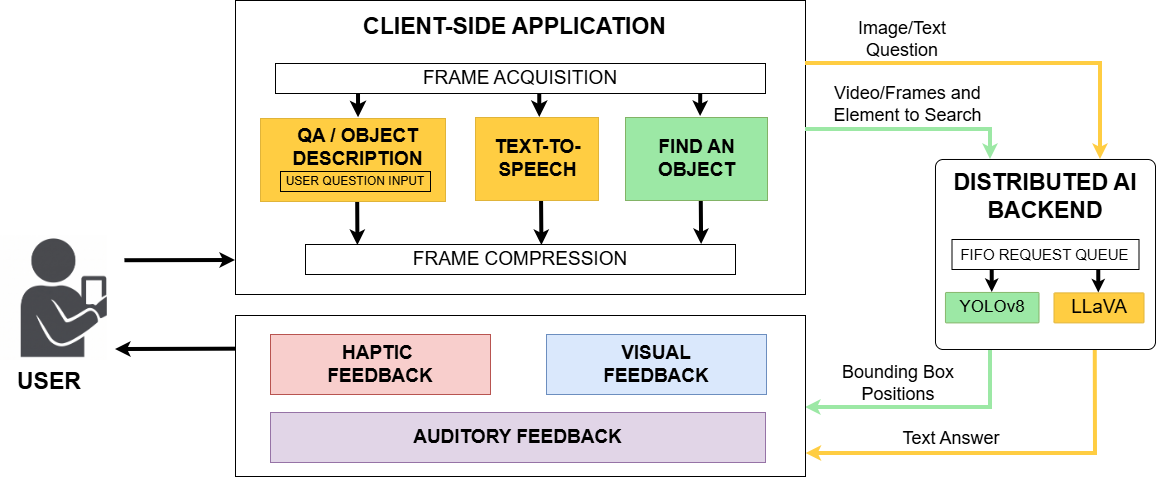}
    \caption{AIDEN’s distributed architecture overview. The orange path shows the LLaVA-based question-answering flow, while the green path shows the YOLOv8-based object detection flow.}
    \label{fig:arch}
\end{figure*}
All functionalities in AIDEN share a unified and consistent layout: approximately 45\% of the screen is dedicated to the camera preview, while the remaining space hosts control buttons and displays the system’s output. This design ensures a seamless and intuitive experience, enabling users to easily capture or select images, access system responses, and navigate through the application.

Given the computational complexity of the algorithms used, such as deep learning–based object detection and VQA, executing all functionalities directly on a mobile device would demand high-end hardware and lead to reduced accuracy or degraded performance. To address this challenge and broaden accessibility across devices, AIDEN adopts a distributed interconnection architecture.

In this architecture, shown in Figure~\ref{fig:arch}, computationally intensive models such as YOLO and a VLM, specifically LLaVA, are executed on a centralized server. The mobile application is responsible solely for capturing input and presenting the output. This setup significantly reduces the hardware requirements on the user's side and simplifies updates to the core models. However, it does require an active internet connection (WiFi or mobile data), which is common among most users. When a user interacts with AIDEN, by selecting a function and capturing an image, the data is sent to the remote server, processed in a first-in-first-out manner, and the result is sent back to the device, where it is presented to the user using visual, audio, or haptic feedback.

{
The centralized server provides the necessary throughput for real-time and near-real-time AI inference, operating on Ubuntu 18.04 LTS with an Intel i7-8700 CPU and 32 GB of RAM. A dual-GPU setup is employed to parallelize distinct AI tasks: an NVIDIA GTX 1080Ti is dedicated to YOLOv8 processing to optimize low-latency spatial guidance for the "Find an Object" module, while an NVIDIA A40 is reserved for LLaVA-based VQA and OCR tasks that require higher VRAM for Vision-Language Model (VLM) inference. This hardware configuration ensures that the distributed architecture can handle high-performance models on resource-constrained mobile devices.
Communication between the smartphone client and the server occurs over 4G LTE or WiFi networks, with data packets handled through a First-In-First-Out (FIFO) queuing model~\cite{knuth1997art}. To maintain system responsiveness, the "Find an Object" functionality is prioritized to minimize round-trip time (RTT), achieving an average processing speed of 1.96 frames per second on the mobile device. For semantic tasks, the system utilizes specialized AI agents that trigger the LLaVA model using specific prompts, such as "Transcribe the text present in this image" for OCR. The resulting structured data is then streamed back to the device and rendered for the user via the Capacitor ScreenReader plugin.}

One of the key features supported by this architecture is OCR. This option allows users to interpret printed text from various surfaces. To perform this task, the system leverages a specialized AI Agent (based on the LLaVA model). This agent, which is configured with a specific role and prompt, uses the image to provide an answer to the user in the OCR option, producing more accurate textual transcriptions from image inputs.

The OCR process involves capturing or uploading an image that contains printed or digital text. The image is then sent to the server, where it is processed by LLaVA using the prompt: \textit{"Transcribe the text present in this image."} The output is a structured textual representation of the detected content.

\begin{figure}[h]
    \centering

    \subfloat[Legumbres PEDRO ALUBIAS beans.]{%
        \includegraphics[width=0.22\textwidth]{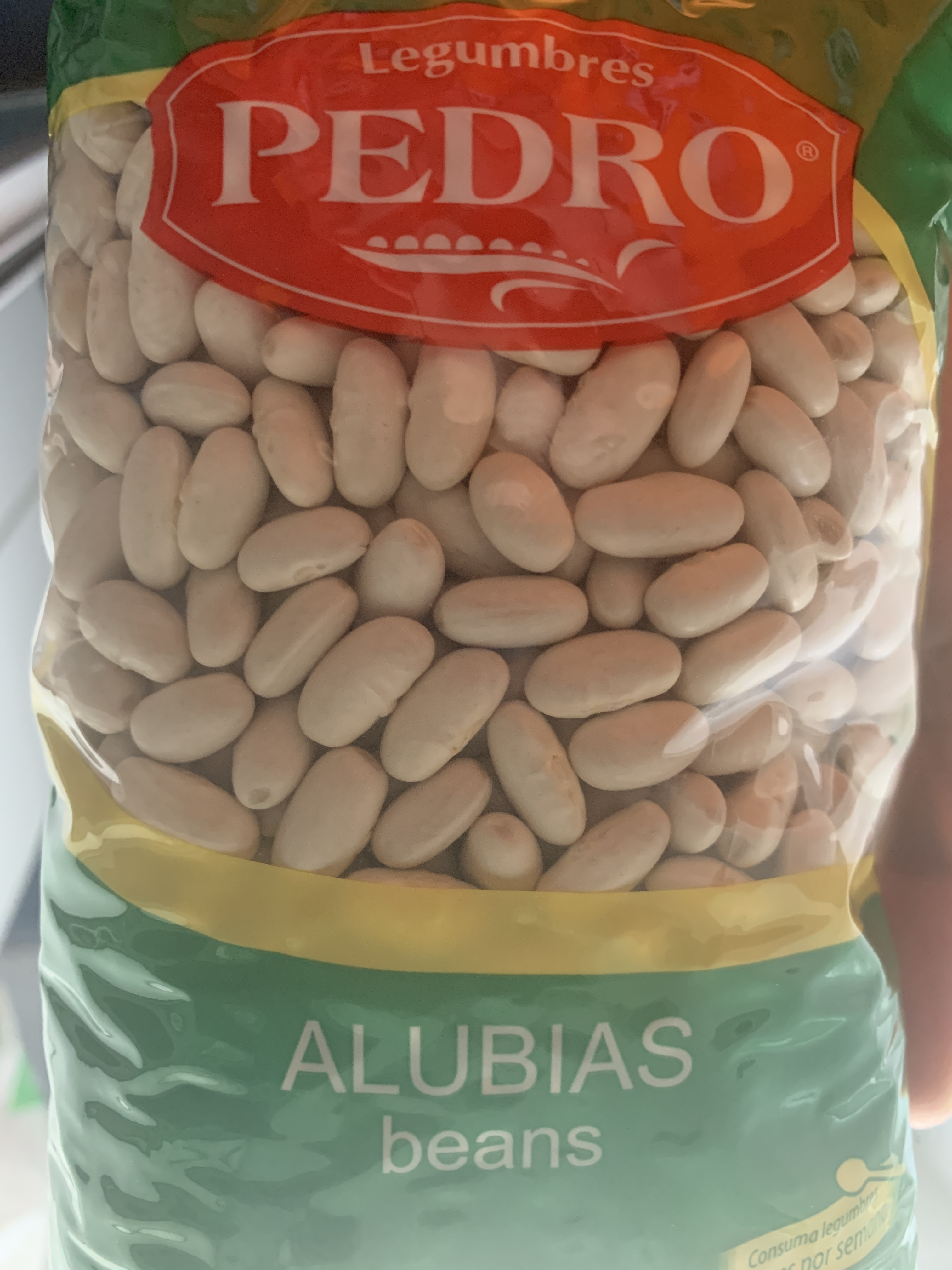}
    }
    \hfill
    \subfloat[...Nolotil 575 mg capsulas duras Metamizol...]{%
        \includegraphics[width=0.22\textwidth]{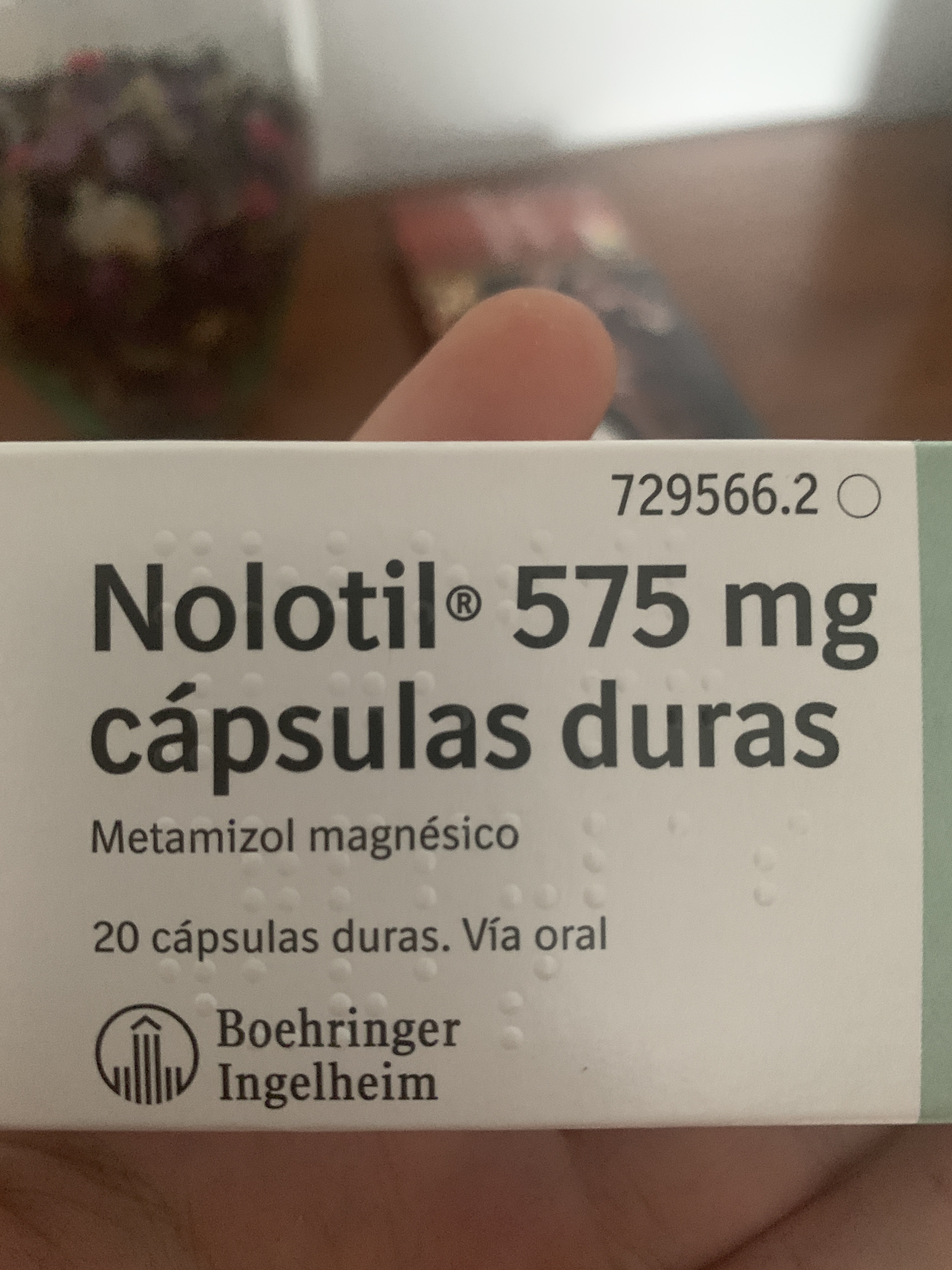}
    }
    \hfill
    \subfloat[A se vi Suavizante Azul pres ccr intense.]{%
        \includegraphics[width=0.22\textwidth]{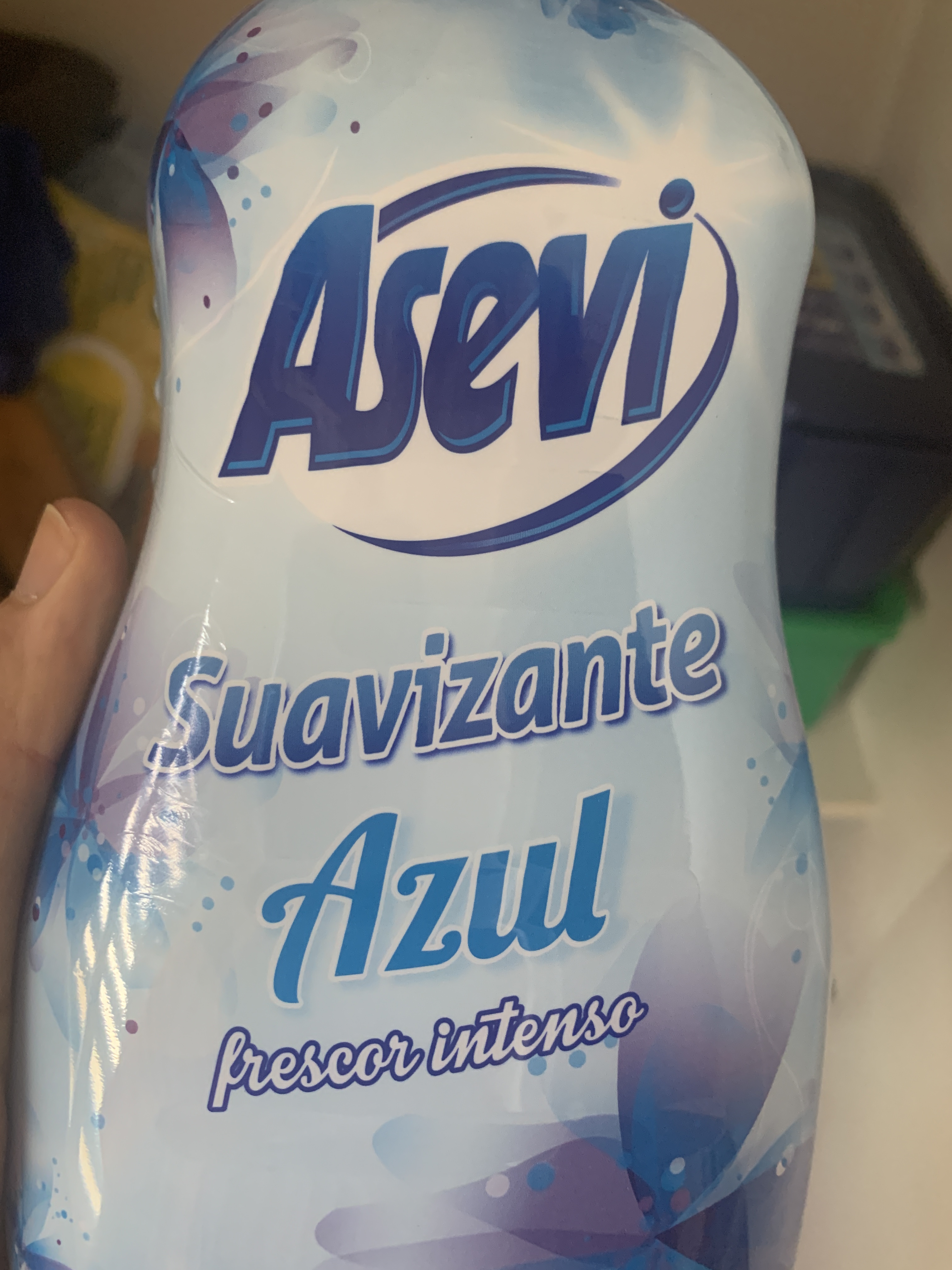}
    }
    \hfill
    \subfloat[CARRER DE L'ARTISTA FOGUERER.]{%
        \includegraphics[width=0.22\textwidth]{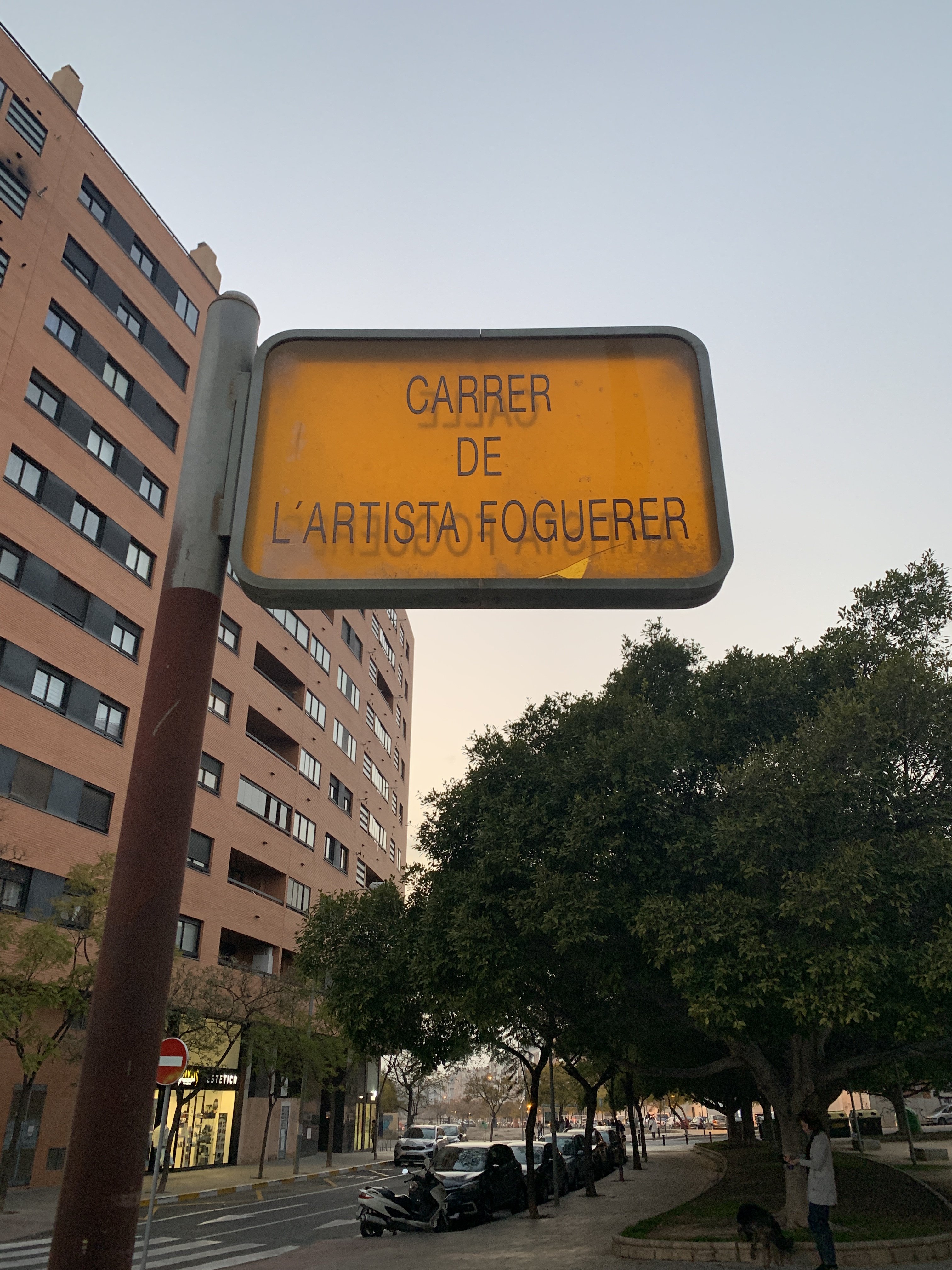}
    }

    \caption{Random samples of input images and retrieved text returned by the OCR feature.}
    \label{fig:ocr}
\end{figure}

LLaVA achieves high transcription accuracy under standard conditions, successfully interpreting a high percentage of characters in images with clearly printed text. While its performance slightly degrades with blurred or distorted inputs, it still produces useful results. Moreover, it supports multilingual recognition. Representative examples of OCR output are shown in Figure~\ref{fig:ocr}.

As mentioned, another feature supported by AIDEN is the ``Find an Object'', which helps users locate specific items within their environment. After selecting the ``Find an Object'' option, the user verbally specifies the item to search for (e.g., ``backpack'') and moves the phone to scan the surrounding area. If the object is detected, the system provides real-time voice instructions to help the user center it in the camera view. Once centered, the user can move in the direction of the phone to retrieve the item.

\begin{figure}[h]
    \centering
    \subfloat[]{%
        \includegraphics[width=0.32\textwidth]{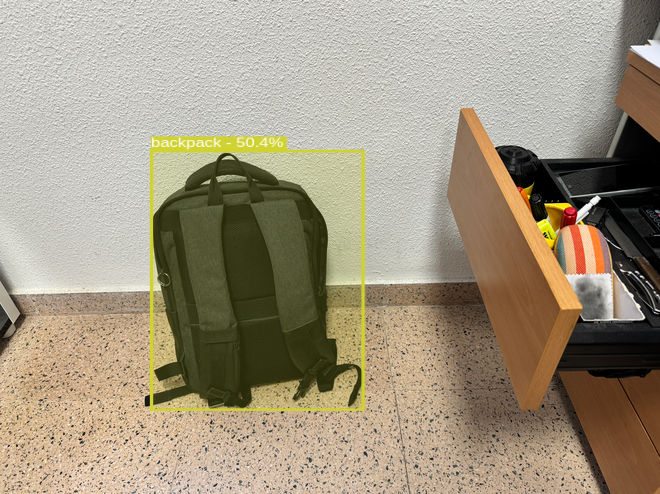}
    }
    \hfill
    \subfloat[]{%
        \includegraphics[width=0.32\textwidth]{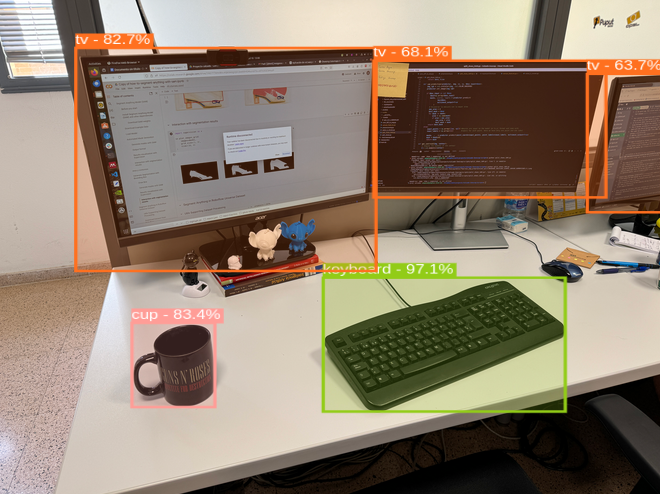}
    }
    \hfill
    \subfloat[]{%
        \includegraphics[width=0.32\textwidth]{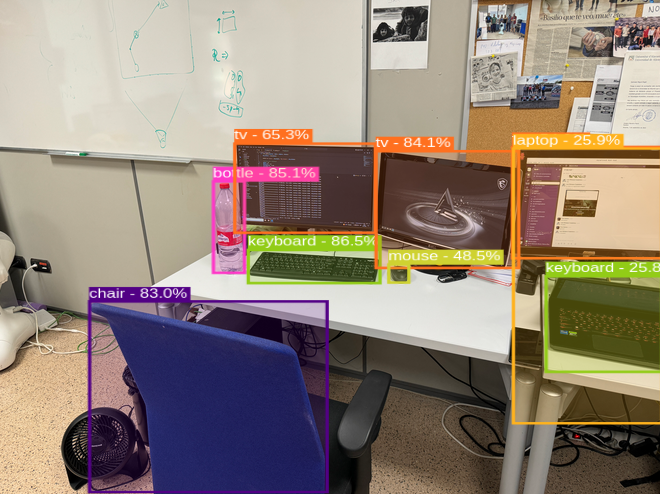}
    }
    \caption{Actual detections of the ``Find an Object'' method in an office environment.}
    \label{fig:yolo}
\end{figure}

This functionality is powered by YOLOv8~\cite{Jocher_Ultralytics_YOLO_2023}, a state-of-the-art object detection model capable of identifying 80 different categories, such as bottles, glasses, and backpacks. To address user needs, we extended the model to include an additional class for ``door'', a frequently requested element not available in the standard model. Examples of detections produced by this feature are shown in Figure~\ref{fig:yolo}.

\subsection{Experimental Setup and Procedure}
Two distinct evaluations were conducted to validate the AIDEN system: a quantitative assessment of runtime performance and a qualitative user study focused on perceived usability and technology acceptance.

\subsubsection{Apparatus and Technical Setup}
For the runtime evaluation and during the user tests, AIDEN was deployed on an LG G6 smartphone (Qualcomm Snapdragon 821, 4~GB RAM, 5.7-inch QHD+ display) connected to a 4G LTE network. To ensure consistent processing capabilities, computationally intensive processing was offloaded to a remote server running Ubuntu 18.04 LTS, equipped with an Intel i7-8700 CPU, 32~GB RAM, and dual GPUs (NVIDIA GTX 1080Ti for YOLO processing and NVIDIA A40 for LLaVA-based tasks).

\subsubsection{User Study Procedure}
The user evaluation took place in a controlled laboratory environment with constant lighting conditions and minimized background noise to optimize the performance of computer vision models and audio feedback. The session for each participant lasted approximately 25 minutes and was divided into three distinct phases:

\paragraph{Phase 1: Training and Familiarization}
Prior to data collection, participants underwent a tutorial session. An instructor guided them through the application's interface, allowing them to explore tactile boundaries and haptic feedback patterns without recording performance metrics. This phase mitigated the "learning curve effect", ensuring that subsequent measurements reflected system usability rather than novelty.

\begin{figure}[H]
    \centering
    \includegraphics[width=0.45\textwidth]{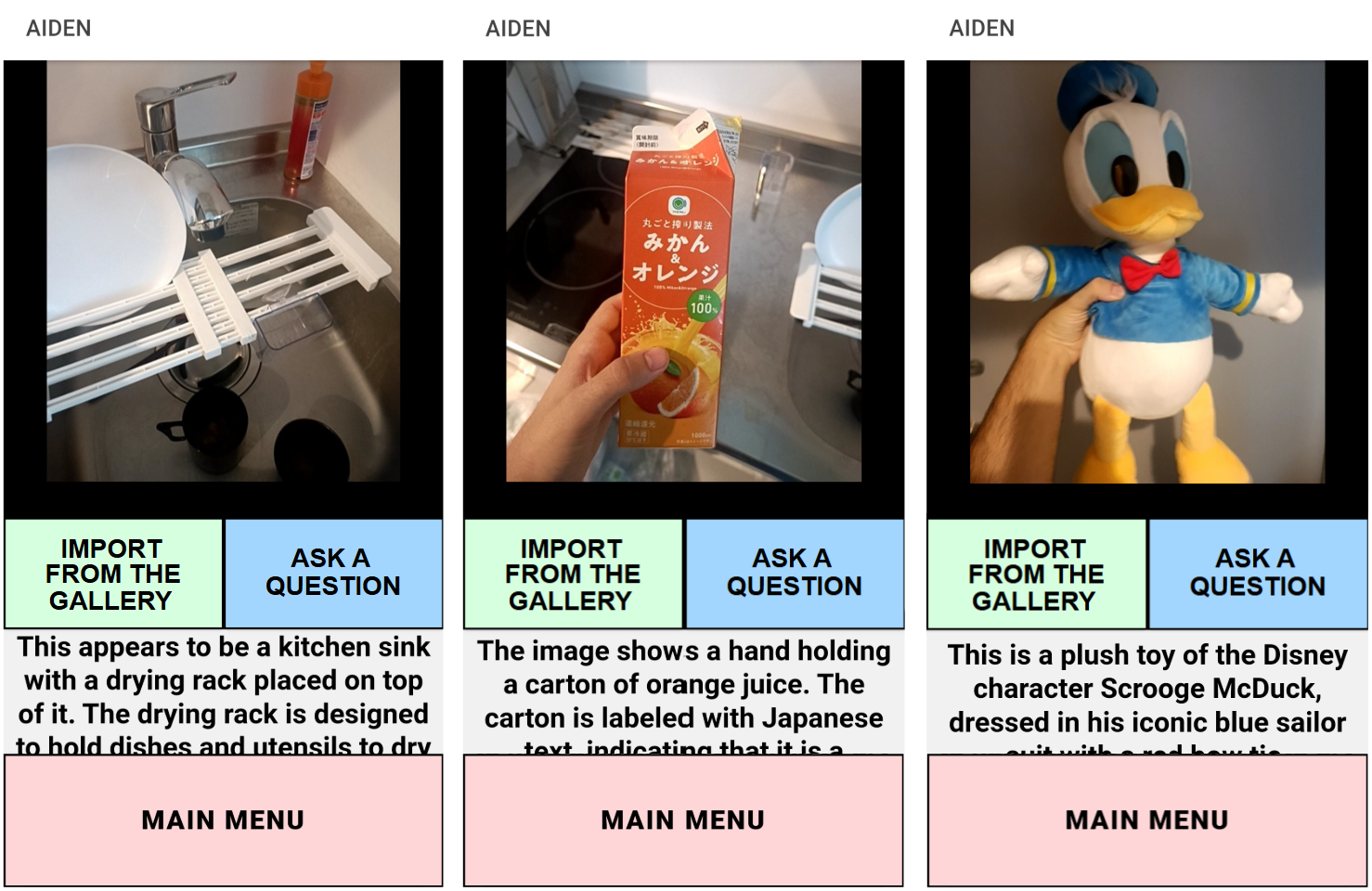}
    \caption{Examples of input images and corresponding answers returned by the Object Description algorithm during the evaluation.}
    \label{fig:questans}
\end{figure}

\paragraph{Phase 2: Task Execution}
Participants performed three high-priority daily living tasks using a counterbalanced order to avoid sequence effects. These tasks evaluated the full spectrum of AIDEN's functionalities:

\begin{itemize}
    \item \textbf{Task A (object retrieval):} 
    Participants stood in front of a table containing clutter and were instructed to locate a specific target item using the spatial guidance features (Section \ref{sec:spatial_guidance}).
    
    \item \textbf{Task B (information access):} 
    Participants were handed a printed document representing a real-world scenario. They used the OCR function to read the text aloud and identify specific information (e.g., expiration date).
    
    \item \textbf{Task C (scene understanding - VQA):} 
    Participants used the LLaVA-based Object Description feature. They pointed the camera at the room to capture an image, which the system processed using the fixed prompt: \textit{“What is this? Provide the answer as summarized as possible.”} Users could also ask follow-up natural language questions (e.g., "Is the door open?"). The generated audio descriptions and their correspondence to the visual input are illustrated in Figure~\ref{fig:questans}.
\end{itemize}

\paragraph{Phase 3: Data Collection and Success Criteria}
During task execution, we employed a Think-Aloud Protocol~\cite{ericsson2017protocol}, encouraging participants to verbalize their thoughts and frustrations. A task was considered "Successful" if the goal was achieved without physical intervention.

Upon completion, participants assessed their subjective perception using the TAM. The questionnaire consisted of 16 items administered via Google Forms, with responses collected on a 5-point Likert scale ~\cite{joshi2015likert}.

\section{Data Preparation and Analysis}
\label{secc:data_prep}
To evaluate both the technical performance and user acceptance of AIDEN, two complementary experiments were conducted: one focused on system runtime and the other on user perception using the TAM.

For the runtime evaluation, each of the system's core functionalities—``Object Description'' with VQA, ``Text-To-Speech'' and ``Find an Object''—was executed 20 times. During each trial, two timing measurements were recorded: the processing time on the remote server, and the total response time as observed on the smartphone, which included image transmission and result retrieval. These data were used to calculate the mean and standard deviation for each functionality, providing insight into the system's consistency and responsiveness under real-world usage conditions.

To facilitate the interpretation of the Likert scores for a broad audience, mean values were mapped to qualitative descriptors inspired by the adjective ratings proposed by Bangor et al.~\cite{bangor2009determining}. Since Perceived ease-of-use (PEOU) in this context serves as a proxy for usability, we adapted their ranges to 5-point Likert scale to qualitatively categorize user satisfaction levels: Very Poor (1.0--1.25), Poor (1.25--2.5), Acceptable (2.5--3.5), Good (3.5--4.0), Excellent (4.0--4.5), and Best (4.5--5.0).

Descriptive statistics, including means and score ranges for each question, were calculated to summarize the overall perception of the system and to highlight particularly higher- and lower-rated areas.

\section{Results}\label{sec:procedure}

The results of the runtime evaluation are presented in Table~\ref{tab:runtime}. Both the ``Object Description'' and ``Text-To-Speech'' functionalities exhibited similar server-side runtimes, as they rely on the same backend infrastructure and models. Slight differences in processing time were observed due to variations in the length and complexity of the generated outputs. In contrast, the ``Find an Object'' functionality, designed for low-latency operation, delivered significantly faster results and achieved an average processing speed of 1.96 frames per second on the smartphone.

\begin{table}[H]
\centering
\renewcommand{\arraystretch}{1.3} 

\begin{tabular}{l c c} 
\hline
\textbf{Functionality} & \textbf{Server (s)} & \textbf{Smartphone (s)} \\ 
\hline
Object Description \& QA       & $7.34 \pm 1.10$ & $10.10 \pm 1.29$ \\
Optical Character Recognition & $7.09 \pm 2.57$ & $9.54 \pm 2.94$ \\
Find an Object (1 req.)        & $0.23 \pm 0.18$ & $0.51 \pm 0.22$ \\ 
\hline
\end{tabular}

\caption{Runtime of the different functionalities AIDEN provides. Values represent mean $\pm$ standard deviation.}
\label{tab:runtime}
\end{table}

\begin{figure*}[t!]
    \centering
    \includegraphics[width=0.9\textwidth]{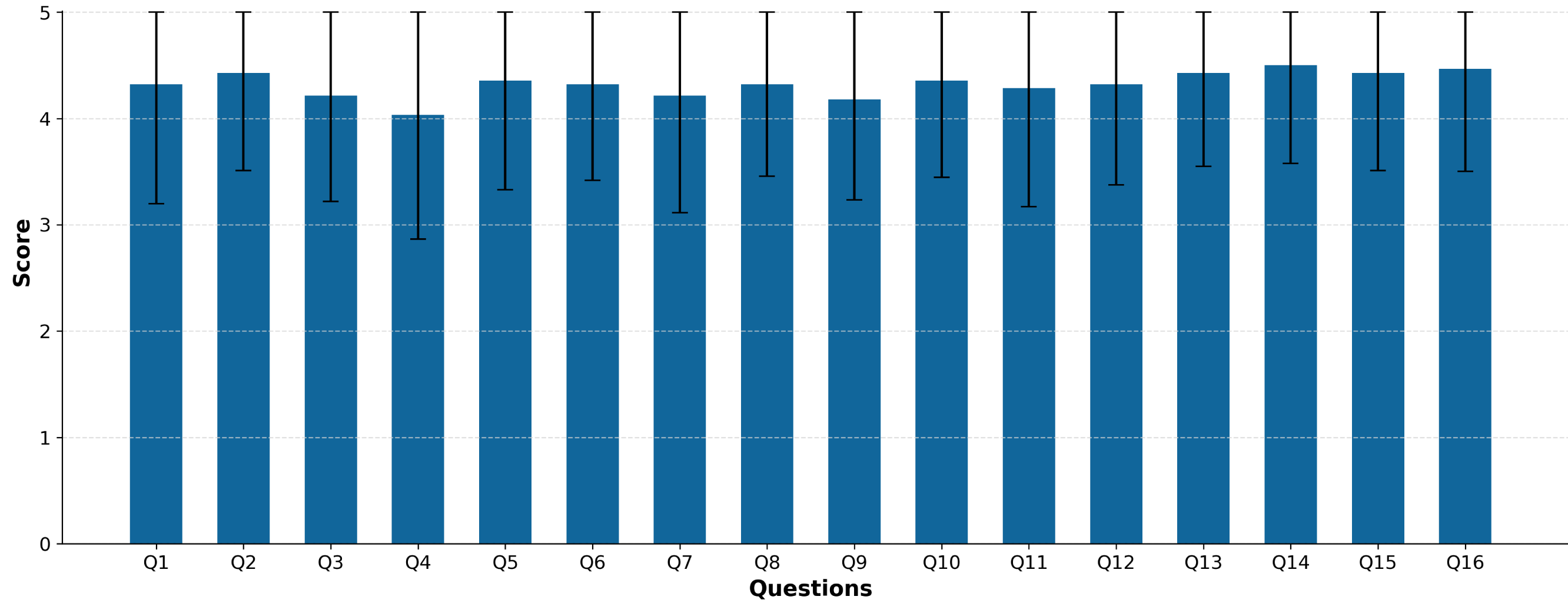}
    \caption{Average score and standard deviation per question obtained by the 16-item questionnaire (Range: 1 = Very Poor, 5 = Best).}
    \label{fig:tam_plot}
\end{figure*}

Regarding user feedback, the TAM questionnaire results are summarized in Figure~\ref{fig:tam_plot}. Overall, participants gave generally high ratings to AIDEN, with average ratings falling between the "Excellent" and "Best" categories for most questions. The highest mean score was observed for participants’ willingness to incorporate the final version of the system into their daily routines (Q14), followed by the overall positive evaluation of the system (Q16). High ratings were also obtained for participants’ intention to use the system in real-world settings and to recommend it to others (Q13 and Q15), suggesting favorable acceptance and interest in future use.

\subsection{Qualitative Observations and User Feedback}
\label{sec:qualitative_results}

To complement the quantitative metrics and TAM scores, we conducted a thematic analysis of the participants' verbal feedback and the behavioral patterns observed during the sessions. This triangulation provides a deeper insight into the practical challenges of deploying AI assistants for visually impaired individuals.

\subsubsection{Behavioral Observations}
During the execution of tasks, the researchers noted specific interaction patterns that influenced the system's performance:

\begin{itemize}
    \item \textbf{Learning curve:}
    {
    Regarding the ``Find an Object'' function, users initially struggled to interpret the varying vibration frequencies. However, after the third attempt in the training phase, participants displayed a smoother scanning motion, suggesting that the continuous haptic guidance scheme is effective but requires a short adaptation period.
    }
    
    \item \textbf{Distance estimation:} 
    In the OCR task (Task B), some participants held the document too close to the lens. Conversely, others held the document too far away or positioned it such that the capture encompassed a broad area of unnecessary text.
    
\end{itemize}

\subsubsection{User Feedback and Perceived Utility}
Participants were encouraged to express their thoughts using a Think-Aloud protocol. The feedback was largely positive regarding autonomy, though constructive criticism was raised regarding physical ergonomics.

\begin{itemize}
    \item \textbf{On autonomy and speed:} 
    Participants valued the response time of the AI models. Participant [P2] remarked: \textit{``I am used to waiting much longer with cloud-based apps. Having the answer instantly makes me feel more confident to keep walking.''} Similarly, Participant [P5] noted: \textit{``The ability to find my keys without calling my son is a huge relief for me.''}
    
    \item \textbf{On interface navigation:}
    While the semantic navigation was praised, some users suggested improvements. Participant [P4] stated: \textit{``The buttons are accessible, but the space between the buttons must be removed.''}
\end{itemize}

These qualitative insights align with the TAM results, suggesting that while the functional utility is high, the physical interaction with the device remains the primary bottleneck for non-visual users.

\subsection{Evaluation Framework}
\label{sec:tam_justification}

To strictly assess the system's potential for real-world adoption, we employed the TAM. We selected TAM over pure usability metrics because, while it captures the perceived usability through the PEOU construct, it specifically correlates this factor with the user's Behavioral Intention to Use (BI). Given the pilot nature of this study, BI is analyzed as the primary predictor of future adoption, serving as a proxy for ``Actual System Use'' which requires longitudinal data not yet available in this development phase:

\begin{enumerate}
    \item \textbf{PU:} defined as the degree to which the user believes that AIDEN enhances their autonomy in daily tasks (e.g., reading without assistance).
    \item \textbf{PEOU:} defined as the degree to which using the system is free of physical and cognitive effort. In the context of visual impairment, PEOU is directly linked to the efficacy of the non-visual interface (haptics and audio).
\end{enumerate}

\subsection{User Acceptance and Usability Analysis}
{To assess} user acceptance, participant responses were analyzed using the TAM. Given the sample size ($N=28$), we adopted non-parametric {statistical methods}. We employed bootstrapping (with $R=2000$ resamples) to generate bias-corrected 95\% confidence intervals. The detailed descriptive statistics for each construct and item are presented in Table~4.

Reliability analysis demonstrated  internal consistency. Cronbach's $\alpha$ values were 0.82 for PU and an exceptional 0.93 for PEOU, both well above the recommended threshold of 0.70~\cite{hair2009multivariate}.

Regarding the balance between constructs, the Wilcoxon Signed-Rank Test~\cite{woolson2007wilcoxon} revealed no significant divergence between the system's utility and its ease of use ($Z \approx 0.77$, $p = 0.4429$). This suggests that, in this sample, neither construct was rated systematically higher than the other. However, this result should be interpreted cautiously: it indicates an absence of evidence for a difference, rather than evidence of equivalence between constructs.

\begin{table*}[htbp]
\centering
\label{tab:tam_robust}
\begin{tabular}{l l c c c c c}
\hline
\textbf{Construct} & \textbf{Item Description} & \textbf{Mdn} & \textbf{IQR} & \textbf{Mean} & \textbf{SD} & \textbf{95\% CI} \\
\hline
\multicolumn{7}{l}{\textit{\textbf{PU (Perceived Usefulness)}}} \\
 & PU1: Useful for daily tasks & 5.0 & 1.0 & 4.36 & 1.03 & [3.89, 4.68] \\
 & PU2:  Information resource & 5.0 & 1.0 & 4.32 & 0.90 & [3.93, 4.61] \\
 & PU3: Effective for personal independence & 5.0 & 1.0 & 4.21 & 1.10 & [3.75, 4.54] \\
 & PU4:  Satisfies specific need & 5.0 & 1.0 & 4.32 & 0.94 & [3.93, 4.61] \\
 & \textit{\textbf{Construct Aggregate}} & \textbf{4.50} & \textbf{0.81} & \textbf{4.30} & \textbf{0.90} & \textbf{[3.90, 4.59]} \\
\hline
\multicolumn{7}{l}{\textit{\textbf{PEOU (Perceived Ease of Use)}}} \\
 & PEOU1:  System is intuitive & 5.0 & 1.0 & 4.32 & 1.12 & [3.82, 4.68] \\
 & PEOU2:  Autonomy in use & 5.0 & 1.0 & 4.43 & 0.92 & [4.04, 4.71] \\
 & PEOU3: Options are easy/clear & 5.0 & 1.0 & 4.29 & 1.12 & [3.82, 4.64] \\
 & \textit{\textbf{Construct Aggregate}} & \textbf{4.83} & \textbf{0.75} & \textbf{4.35} & \textbf{1.00} & \textbf{[3.93, 4.65]} \\
\hline
\multicolumn{7}{l}{\textit{\textbf{ATT (Attitude)}}} \\
 & ATT1: Users will enjoy using it & 5.0 & 1.25 & 4.21 & 0.99 & [3.79, 4.54] \\
 & ATT2: Motivating system & 4.0 & 1.25 & 4.04 & 1.17 & [3.57, 4.43] \\
 & ATT3: Well thought out & 4.5 & 1.0 & 4.32 & 0.86 & [3.93, 4.57] \\
 & ATT4: Attractive design & 4.0 & 1.0 & 4.18 & 0.94 & [3.82, 4.50] \\
 & ATT5: Attractive options & 5.0 & 1.0 & 4.36 & 0.91 & [3.96, 4.64] \\
 & ATT6:  Positive overall rating & 5.0 & 1.0 & 4.46 & 0.96 & [4.00, 4.75] \\
 & \textit{\textbf{Construct Aggregate}} & \textbf{4.50} & \textbf{0.67} & \textbf{4.26} & \textbf{0.82} & \textbf{[3.90, 4.51]} \\
\hline
\multicolumn{7}{l}{\textit{\textbf{BI (Behavioral Intention)}}} \\
 & BI1: I would like to use it & 5.0 & 1.0 & 4.43 & 0.88 & [4.00, 4.68] \\
 & BI2: Introduce final version & 5.0 & 1.0 & 4.50 & 0.92 & [4.07, 4.79] \\
 & BI3: I would recommend it & 5.0 & 1.0 & 4.43 & 0.92 & [4.00, 4.71] \\
 & \textit{\textbf{Construct Aggregate}} & \textbf{4.83} & \textbf{1.00} & \textbf{4.45} & \textbf{0.81} & \textbf{[4.08, 4.70]} \\
\hline
\end{tabular}

\label{sec:tam_results}
\caption{Descriptive statistics of the TAM questionnaire including robust estimators. CI represents the 95\% bootstrapped confidence interval. Mdn = Median; IQR = Interquartile Range; SD = Standard Deviation. Range: 1 = Very Poor, 5 = Best.}
\end{table*}

\section{Discussion}
\label{sec:discussion}

The empirical results indicate that the proposed distributed architecture integrates ``Object Description'' with VQA, ``Text-To-Speech'' and ``Find an Object'' into a coherent interaction paradigm. {The results suggest that the integration logic may have contributed to the observed performance and user ratings. The high ratings in PEOU ($\mu=4.35$) and ATT ($\mu=4.26$) suggest that the interaction design may reduce cognitive burden, supporting its potential use in daily contexts.} Notably, the ``Find an Object'' module achieved near real-time latency, delivering fluid feedback that may contribute to a smoother interaction experience.

Beyond the standard constructs of the TAM, our analysis provided additional insights regarding the determinants of adoption. While both PU ($\mu=4.30$) and PEOU ($\mu=4.35$) were rated highly, Spearman correlation analysis revealed that BI was most strongly correlated with the Attitude Towards Using (ATT) construct ($\rho = 0.805$, $p < 0.001$).

This finding suggests that, for this specific class of AT, users' intention to adopt the system ($\mu=4.45$, 95\% CI $[4.08, 4.70]$) was more strongly associated with intrinsic motivation and perceived design quality, rather than by functional utility alone. Participants characterized the system not merely as useful but as engaging, a distinct quality that may have contributed to their willingness to recommend and integrate it into their daily routines.

From a user-centered perspective, the high ratings for PU suggest that AIDEN appears to address user-reported needs in this sample. Qualitative feedback described the application as intuitive, a result largely attributed to the rigorous emphasis on usability via multimodal feedback, specifically the integration of voice guidance and haptic signals. This underscores the value of AIDEN not just as a technical demonstration, but as a practical tool aligned with foundational Human–Computer Interaction principles for accessibility.

One finding of this study is the observed structural relationship between acceptance factors. While traditional TAM literature typically identifies PU as the strongest direct predictor of BI~\cite{huang2019examining}, our analysis suggests that ATT appears to be the factor most strongly associated with intention to use AIDEN ($\rho=0.805, p<0.001$). This elevated role of Attitude suggests a possible shift from purely functional drivers to intrinsic motivators:

\begin{enumerate}
    \item \textbf{Intrinsic motivation vs. performance expectancy:} Unlike studies applying UTAUT~\cite{venkatesh2012consumer}, to visually impaired populations, where performance expectancy typically dictates adoption~\cite{moon2022factors}, the strong correlation between ATT and BI in our study indicates that the quality of the experience may play an important role. This contrasts with findings in other contexts, such as mobile payments for visually impaired individuals in India, where Ease of Use was the primary determinant~\cite{parvathy2022adoption}. In the case of AIDEN, user adoption may be driven not solely by the system's utility (what it does), but by the subjective satisfaction of the interaction (how users feel while using it).  This apparent influence of ATT could arise from the contrast between the frustration experienced in manual, unsupported tasks and the relative relief provided by the application.
    
    \item \textbf{Experience economy in assistive tech:}
    Prior work has integrated an Experience Economy–based construct into an extended TAM for hotel mobile applications, showing that experiential factors can strongly predict behavioral intention alongside perceived usefulness~\cite{huang2019examining}. 
    Complementing this, research on high-tech assistive technologies for students with visual impairment highlights the central role of intrinsic factors such as self-confidence, technophobia and attitudes in determining whether these technologies are actually used~\cite{kisanga2025enhancing}. 
    Building on our findings, we therefore hypothesize that, for assistive technologies, design choices that foster users' trust and confidence (e.g., through clear and reassuring feedback such as haptics) may enhance users’ attitudes toward using the system and may be as relevant to future adoption as purely objective efficiency metrics.

\end{enumerate}

Although this study focused specifically on visual impairments, the architectural design of AIDEN strictly adheres to the native accessibility standards of the mobile operating system. This adherence ensures theoretical compatibility with system-level assistive technologies. Consequently, users with limited fine motor control can interact with AIDEN using external input devices or voice commands, bypassing the need for touch gestures. This inherent compatibility extends the system's potential impact beyond the demographic examined in this preliminary study.

\section{Limitations and Conclusion}
\label{sec:limitations_conclusions}

Interpreting the findings presented above requires careful consideration of the study's constraints to contextualize our contribution and outline the roadmap for future research. Regarding sample size and demographics, our user study involved a cohort of 28 participants ($N=28$). While this sample size is consistent with usability testing standards for specialized populations to identify major design flaws, it limits the statistical power required to generalize the findings to the broader visually impaired community. Consequently, {our results provide a preliminary validation of the system integration and interaction design, establishing a foundation for future research into user-centered assistive frameworks.} Future iterations will aim to recruit larger, more demographically diverse cohorts. Furthermore, the evaluation was conducted in a controlled laboratory setting. While this ensured participant safety and variable consistency, it does not fully reflect the stochastic nature of ``in-the-wild'' {deployment conditions}. Factors such as dynamic outdoor lighting, acoustic interference in crowded spaces, {background distractions, scene clutter,} or intermittent network connectivity were not stress-tested in this initial phase. {These factors may affect both perception performance and user interaction, particularly in time-sensitive assistive scenarios where delayed, incomplete, or unstable feedback may reduce usability and user trust. As a result, the present findings should be interpreted as an initial validation under controlled conditions rather than as evidence of full robustness in everyday real-world use.}

A significant technical constraint is the reliance on server-side inference. Although this distributed architecture enables the use of high-performance models like LLaVA and YOLOv8 on resource-constrained devices, it introduces a mandatory dependence on an active internet connection. In ``in-the-wild'' environments, factors such as intermittent network coverage{, unstable WiFi availability, } or 4G/LTE latency {and jitter} could compromise the low-latency nature of the ``Find an Object'' module. {This issue is especially relevant for functions} {relying on high-frequency closed-loop feedback, since} {delays in transmission, inference, or server response} {or inconsistent signal quality} {may reduce the usefulness of the guidance provided} {at the moment it is needed.} This reliance also poses a potential bottleneck for autonomy in remote or poorly connected areas{, and may limit the reliability of the system outside well-connected urban settings}. Consequently, our findings establish a foundation for future research into Edge AI integration, aiming to perform critical tasks locally to enhance safety and decrease server dependency.

Turning to technical and design constraints, a critical challenge inherent to integrating VLMs is the potential for hallucinations. In the context of AT for visually impaired users, this becomes a safety-critical issue. If the system incorrectly identifies a clear path or a non-existent object without providing a confidence metric, it risks violating the robustness and understandability principles of the WCAG. The current iteration of AIDEN, like existing commercial applications, does not yet feature a real-time uncertainty quantification mechanism to alert users when the AI's confidence is low, posing a risk that users might place undue trust in erroneous descriptions. {This limitation is particularly important under real-world deployment conditions, where ambiguous scenes, poor illumination, motion blur, or partial occlusions may increase the likelihood of unreliable outputs. In such cases, a plausible but incorrect description may be more harmful than no description at all. Therefore, mitigating hallucinations is a necessary step for safe real-world deployment.} User safety is intrinsically linked to interaction design. The current reliance on complex touch gestures risks excluding users with accompanying motor impairments. Strictly adhering to WCAG 2.1 Criterion 2.5.1 (Pointer Gestures), the application must ensure that all functions are operable via simple inputs and provide touch targets of at least $44 \times 44$ pixels. The current prototype requires further refinement to offer redundant, large tactile controls for robust accessibility.

Building on these identified limitations, future research will focus on four key areas. First, we plan to incorporate controlled comparative evaluations to better isolate the contribution of AIDEN’s multimodal design. In particular, future studies should include baseline conditions, such as an audio-only version of the system, to assess the incremental benefit of haptic guidance and multimodal feedback. In addition, module-level ablation studies should be conducted by selectively disabling or isolating components such as object detection, OCR, scene description, and Geiger-counter guidance. These comparisons would help disentangle the contribution of individual modules and provide a clearer basis for interpreting usability, confidence, and adoption outcomes. Second, we plan to deploy AIDEN in a longitudinal ``beta testing'' program for daily use over several weeks to assess technology acceptance{, robustness, reliability,} and usability under real-world conditions. Third, to mitigate the risk of hallucinations, future work will implement confidence scoring layers, allowing the system to verbally qualify its descriptions{, warn users when output uncertainty is high, cross-check outputs across modules when possible,} or withhold output if the confidence threshold is not met. {Additional mitigation strategies will explore conservative response policies for safety-critical queries and fallback behaviors that explicitly inform the user when the system cannot provide a reliable answer.} Finally, we aim to integrate Edge AI capabilities to reduce dependency on internet connectivity and expand the ``Find an Object'' feature to support personalized object enrollment and smart home integration.

\section*{Ethical Considerations}
The need for specific ethical clearance was obviated primarily because participant data was rigorously anonymized. This measure was implemented in strict accordance with the GDPR guidelines, ensuring that no Personal Identifiable Information (PII) was stored or shared throughout the study.

\section*{Acknowledgment}

This work was fully supported by Indra and Fundación Universia. The authors also acknowledge the use of Google's Gemini 3 Pro for assistance with the translation and linguistic refinement of this manuscript.

\printbibliography[title={REFERENCES}]

@inproceedings{manduchi2012mobile,
  title={Mobile vision as assistive technology for the blind: An experimental study},
  author={Manduchi, Roberto},
  booktitle={International Conference on Computers for Handicapped Persons},
  pages={9--16},
  year={2012},
  organization={Springer}
}

@book{knuth1997art,
  title={The Art of Computer Programming: Fundamental Algorithms, Volume 1},
  author={Knuth, Donald E},
  year={1997},
  publisher={Addison-Wesley Professional}
}

@article{xia2021haptic,
  title={Haptic feedback interface in navigation systems for visually impaired and blind people: a systematic review},
  author={Xia, Bei and Cho, Youngjun},
  journal={Preprint},
  year={2021}
}

@article{quinn2024shape,
  title={A shape-changing haptic navigation interface for vision impairment},
  author={Quinn, Robert and Murtough, Stephen and de Winton, Henry and Ellis-Frew, Brandon and Zane, Sebastiano and De Sousa, Jonathan and Kempapidis, Theofilos and Gomes, Renata SM and Spiers, Adam J},
  journal={Scientific Reports},
  volume={14},
  number={1},
  pages={29223},
  year={2024},
  publisher={Nature Publishing Group UK London}
}

@article{katzschmann2018safe,
  title={Safe local navigation for visually impaired users with a time-of-flight and haptic feedback device},
  author={Katzschmann, Robert K and Araki, Brandon and Rus, Daniela},
  journal={IEEE Transactions on Neural Systems and Rehabilitation Engineering},
  volume={26},
  number={3},
  pages={583--593},
  year={2018},
  publisher={IEEE}
}

@article{hair2009multivariate,
  title={Multivariate data analysis},
  author={Hair, Joseph F},
  year={2009}
}

@inproceedings{nielsen1993mathematical,
  title={A mathematical model of the finding of usability problems},
  author={Nielsen, Jakob and Landauer, Thomas K},
  booktitle={Proceedings of the INTERACT'93 and CHI'93 conference on Human factors in computing systems},
  pages={206--213},
  year={1993}
}

@article{davis1989technology,
  title={Technology acceptance model},
  author={Davis, Fred D and Bagozzi, Richard P and Warshaw, Paul R},
  journal={J Manag Sci},
  volume={35},
  number={8},
  pages={982--1003},
  year={1989},
  publisher={Springer}
}

@article{matter2017assistive,
  title={Assistive technology in resource-limited environments: a scoping review},
  author={Matter, Rebecca and Harniss, Mark and Oderud, Tone and Borg, Johan and Eide, Arne H},
  journal={Disability and Rehabilitation: Assistive Technology},
  volume={12},
  number={2},
  pages={105--114},
  year={2017},
  publisher={Taylor \& Francis}
}

@incollection{korff1958geiger,
  title={Geiger counters},
  author={Korff, Serge A},
  booktitle={Nuclear Instrumentation II/Instrumentelle Hilfsmittel der Kernphysik II},
  pages={52--85},
  year={1958},
  publisher={Springer}
}

@software{Jocher_Ultralytics_YOLO_2023,
author = {Jocher, Glenn and Chaurasia, Ayush and Qiu, Jing},
license = {AGPL-3.0},
month = jan,
title = {{Ultralytics YOLO}},
url = {https://github.com/ultralytics/ultralytics},
version = {8.0.0},
year = {2023}
}

@misc{BeMyEyes2024,
  author       = {{Be My Eyes}},
  title        = {Be My Eyes: Lend Your Eyes to the Blind},
  year         = {2024},
  url          = {https://www.bemyeyes.com},
  note         = {Accessed: 2025-11-30}
}

@misc{SeeingAI2024,
  author       = {{Seeing AI}},
  title        = {Seeing AI - Talking Camera App for the Blind},
  year         = {2024},
  url          = {https://www.seeingai.com/},
  note         = {Accessed: 2025-11-30}
}

@manual{envision2026,
  title        = {Envision AI},
  author       = {{Envision Technologies B.V.}},
  note         = {Smartphone application},
  url          = {https://www.letsenvision.com/},
}

@article{granquist2021evaluation,
  title={Evaluation and comparison of artificial intelligence vision aids: Orcam myeye 1 and seeing ai},
  author={Granquist, Christina and Sun, Susan Y and Montezuma, Sandra R and Tran, Tu M and Gage, Rachel and Legge, Gordon E},
  journal={Journal of Visual Impairment \& Blindness},
  volume={115},
  number={4},
  pages={277--285},
  year={2021},   
publisher={SAGE Publications Sage CA: Los Angeles, CA}
}

@article{joshi2015likert,
  title={Likert scale: Explored and explained},
  author={Joshi, Ankur and Kale, Saket and Chandel, Satish and Pal, D Kumar},
  journal={British journal of applied science \& technology},
  volume={7},
  number={4},
  pages={396},
  year={2015},
  publisher={Sciencedomain International}
}

@techreport{wcag21,
  type = {W3C Recommendation},
  title = {Web {Content} {Accessibility} {Guidelines} ({WCAG}) 2.1},
  author = {Andrew Kirkpatrick and Joshue O Connor and Alastair Campbell and Michael Cooper},
  year = {2018},
  month = {6},
  institution = {World Wide Web Consortium (W3C)},
  url = {https://www.w3.org/TR/WCAG21/}
}

@article{virzi1992refining,
  title={Refining the test phase of usability evaluation: how many subjects is enough?},
  author={Virzi, Robert A},
  journal={Human factors},
  volume={34},
  number={4},
  pages={457--468},
  year={1992},
  publisher={SAGE Publications Sage CA: Los Angeles, CA}
}

@inproceedings{ahmed2015privacy,
  title={Privacy concerns and behaviors of people with visual impairments},
  author={Ahmed, Tousif and Hoyle, Roberto and Connelly, Kay and Crandall, David and Kapadia, Apu},
  booktitle={Proceedings of the 33rd annual ACM conference on human factors in computing systems},
  pages={3523--3532},
  year={2015}
}

@article{bangor2009determining,
  title={Determining what individual SUS scores mean: Adding an adjective rating scale},
  author={Bangor, Aaron and Kortum, Philip and Miller, James},
  journal={Journal of usability studies},
  volume={4},
  number={3},
  pages={114--123},
  year={2009},
  publisher={Usability Professionals' Association Bloomingdale, IL}
}

@misc{capacitor_screen_reader,
  title = {{Screen Reader Capacitor Plugin API}},
  author = {{Ionic}},
  journal = {Capacitor Documentation},
  howpublished = {\url{https://capacitorjs.com/docs/apis/screen-reader}},
  note = {Accessed: 2023-09-24}
}

@misc{vue_i18n,
  title = {Vue I18n},
  author = {Kawaguchi, Kazuya},
  journal = {Intlify},
  howpublished = {\url{https://vue-i18n.intlify.dev/}},
  note = {Accessed: 2023-09-24}
}

@misc{MetaRayBan,
  author = {{Meta Platforms, Inc.}},
  title = {Ray-Ban Meta Smart Glasses},
  howpublished = {\url{https://www.meta.com/smart-glasses/}},
  year = {2024},
  note = {Accessed: 2025-11-30}
}

@inproceedings{djamasbi2006accessibility,
  author    = {Soussan Djamasbi and Thomas Tullis and Matthew Girouard and Michael Hebner and Jason Krol and Michael Terranova},
  title     = {Web Accessibility for Visually Impaired Users: Extending the Technology Acceptance Model (TAM)},
  booktitle = {Proceedings of AMCIS 2006},
  year      = {2006}
}

@article{bag2016acceptance,
  author  = {Anima Bag and others},
  title   = {Acceptance factors for the use of video call via smartphone by blind people},
  journal = {Technology in Society}, % u otra revista correcta
  year    = {2016}
}

@article{theodorou2024challenges,
  title={Challenges in Acceptance of Smartphone-Based Assistive Technologies: Extending the UTAUT Model for People With Visual Impairments},
  author={Theodorou, Paraskevi and Tsiligkos, Kleomenis and Meliones, Apostolos},
  journal={Journal of Visual Impairment \& Blindness},
  volume={118},
  number={1},
  pages={18--30},
  year={2024},
  publisher={SAGE Publications Sage CA: Los Angeles, CA}
}

@article{kisanga2025enhancing,
  title={Enhancing High-tech Assistive Technology Use for Learning among Students with Visual Impairment in Tanzania’s Higher Education},
  author={Kisanga, Sarah Ezekiel},
  journal={University of Dar es Salaam Library Journal},
  volume={20},
  number={1},
  pages={226--239},
  year={2025}
}

@article{adam2025leveraging,
  title={Leveraging assistive technology for visually impaired people through optimal deep transfer learning based object detection model},
  author={Adam, Mahir Mohammed Sharif and Aljehane, Nojood O and Alzahrani, Mohammed Yahya and Al Zanin, Samah},
  journal={Scientific Reports},
  volume={15},
  number={1},
  pages={30113},
  year={2025},
  publisher={Nature Publishing Group UK London}
}

@article{stevens2005auditory,
  title={Auditory perceptual consolidation in early-onset blindness},
  author={Stevens, Alexander A and Weaver, Kurt},
  journal={Neuropsychologia},
  volume={43},
  number={13},
  pages={1901--1910},
  year={2005},
  publisher={Elsevier}
}

@article{theodorou2019developing,
  title={Developing apps for people with sensory disabilities, and implications for technology acceptance models},
  author={Theodorou, Paraskevi and Meliones, Apostolos},
  journal={Global Journal of Information Technology: Emerging Technologies},
  volume={9},
  number={2},
  pages={33--40},
  year={2019}
}

@article{moon2022factors,
  title={Factors influencing the intention of persons with visual impairment to adopt mobile applications based on the UTAUT model},
  author={Moon, Hyunchang and Cheon, Jongpil and Lee, Jaehoon and Banda, Devender R and Griffin-Shirley, Nora and Ajuwon, Paul M},
  journal={Universal Access in the Information Society},
  volume={21},
  number={1},
  pages={93--107},
  year={2022},
  publisher={Springer}
}

@article{parvathy2022adoption,
  title={Adoption of mobile payment among visually impaired users in Tamil Nadu based on technology acceptance model (TAM)},
  author={Parvathy, V and Durairaj, D},
  journal={International journal of health sciences},
  volume={6},
  number={S3},
  pages={5346--5361},
  year={2022},
  publisher={ScienceScholar}
}

@article{huang2019examining,
  title={Examining an extended technology acceptance model with experience construct on hotel consumers’ adoption of mobile applications},
  author={Huang, Yu-Chih and Chang, Lan Lan and Yu, Chia-Pin and Chen, Joseph},
  journal={Journal of Hospitality Marketing \& Management},
  volume={28},
  number={8},
  pages={957--980},
  year={2019},
  publisher={Taylor \& Francis}
}

@article{woolson2007wilcoxon,
  title={Wilcoxon signed-rank test},
  author={Woolson, Robert F},
  journal={Wiley encyclopedia of clinical trials},
  pages={1--3},
  year={2007},
  publisher={Wiley Online Library}
}

@article{venkatesh2012consumer,
  title={Consumer acceptance and use of information technology: extending the unified theory of acceptance and use of technology},
  author={Venkatesh, Viswanath and Thong, James YL and Xu, Xin},
  journal={MIS quarterly},
  pages={157--178},
  year={2012},
  publisher={JSTOR}
}

@article{ericsson2017protocol,
  title={Protocol analysis},
  author={Ericsson, K Anders},
  journal={A companion to cognitive science},
  pages={425--432},
  year={2017},
  publisher={Wiley Online Library}
}

@inproceedings{zhao2024hearing,
  title={Hearing the World: A Pilot Study Design on Spatial Audio for the Visually Impaired},
  author={Zhao, Xinyan},
  booktitle={Proceedings of the 27th International Academic Mindtrek Conference},
  pages={244--248},
  year={2024}
}

@article{ahmetovic2023enhancing,
  title={Enhancing screen reader intelligibility in noisy environments},
  author={Ahmetovic, Dragan and Galimberti, Gabriele and Avanzini, Federico and Bernareggi, Cristian and Ludovico, Luca Andrea and Presti, Giorgio and Vasco, Gianluca and Mascetti, Sergio},
  journal={IEEE Transactions on Human-Machine Systems},
  volume={53},
  number={4},
  pages={771--780},
  year={2023},
  publisher={IEEE}
}

@article{mrnavi2025,
  title={MR. NAVI: Mixed-Reality Navigation Assistant for the Visually Impaired},
  author={Pfitzer, Nicolas and Zhou, Yifan and Poggensee, Marco and Kurtulus, Defne and Dominguez-Dager, Bessie and Dusmanu, Mihai and Pollefeys, Marc and Bauer, Zuria},
  journal={arXiv preprint arXiv:2506.05369},
  year={2025}
}

\end{document}